\documentclass{article}

\PassOptionsToPackage{numbers, compress}{natbib}

\usepackage[preprint]{neurips_2026}

\usepackage[utf8]{inputenc} %
\usepackage[T1]{fontenc}    %
\usepackage{hyperref}       %
\usepackage{url}            %
\usepackage{booktabs}       %
\usepackage{amsfonts}       %
\usepackage{nicefrac}       %
\usepackage{microtype}      %
\usepackage[table]{xcolor}  %
\usepackage{graphicx}       %
\usepackage{subcaption}     %
\usepackage{wrapfig}        %
\usepackage{amsmath}        %
\usepackage{multirow}      %
\usepackage[capitalise,nameinlink]{cleveref}
\crefname{section}{Sec.\@}{Secs.\@}
\Crefname{section}{Section}{Sections}
\crefname{subsection}{Sec.\@}{Secs.\@}
\Crefname{subsection}{Section}{Sections}
\crefname{figure}{Fig.\@}{Figs.\@}
\Crefname{figure}{Figure}{Figures}
\crefname{table}{Tab.\@}{Tabs.\@}
\Crefname{table}{Table}{Tables}
\crefname{equation}{Eq.\@}{Eqs.\@}
\Crefname{equation}{Equation}{Equations}
\crefformat{equation}{Eq.~#2#1#3}
\Crefformat{equation}{Equation~#2#1#3}
\crefrangeformat{equation}{Eqs.~#3#1#4--#5#2#6}
\Crefrangeformat{equation}{Equations~#3#1#4--#5#2#6}
\crefmultiformat{equation}{Eqs.~#2#1#3}{ and~#2#1#3}{,~#2#1#3}{ and~#2#1#3}
\Crefmultiformat{equation}{Equations~#2#1#3}{ and~#2#1#3}{,~#2#1#3}{ and~#2#1#3}
\crefname{appendix}{Appx.\@}{Appxs.\@}
\Crefname{appendix}{Appendix}{Appendices}
\usepackage{todonotes}

\title{Aligning Latent Geometry for Spherical Flow Matching in Image Generation}

\author{%
  \setcounter{footnote}{1}%
  Tuna Han Salih Meral$^{1}$\thanks{Equal Contribution}\quad Kaan Oktay$^{2}$\footnotemark[\value{footnote}]\quad Hidir Yesiltepe$^{1}$\And Adil Kaan Akan$^{2}$ \quad Pinar Yanardag$^{1}$ \\\\
  $^{1}$Virginia Tech \quad $^{2}$fal
}

\begin{document}

\maketitle
    
\begin{figure*}[!ht]
	\vspace{-15pt}
	\centering
	\begin{subfigure}[t]{0.32\textwidth}
		\centering
		\includegraphics[width=\linewidth]{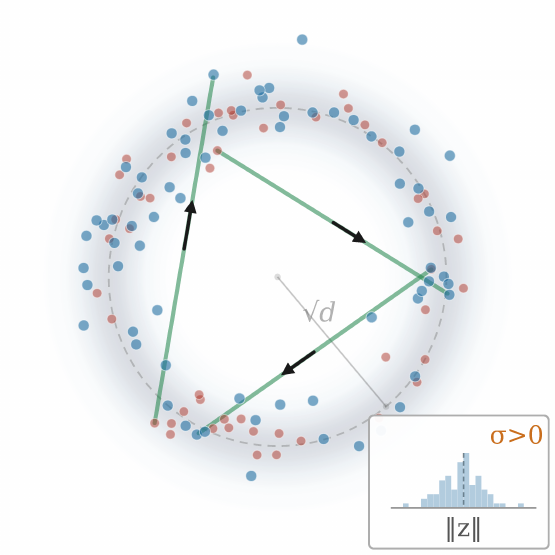}
		\caption{Linear}
		\label{fig:vanilla-linear}
	\end{subfigure}
	\hfill
	\begin{subfigure}[t]{0.32\textwidth}
		\centering
		\includegraphics[width=\linewidth]{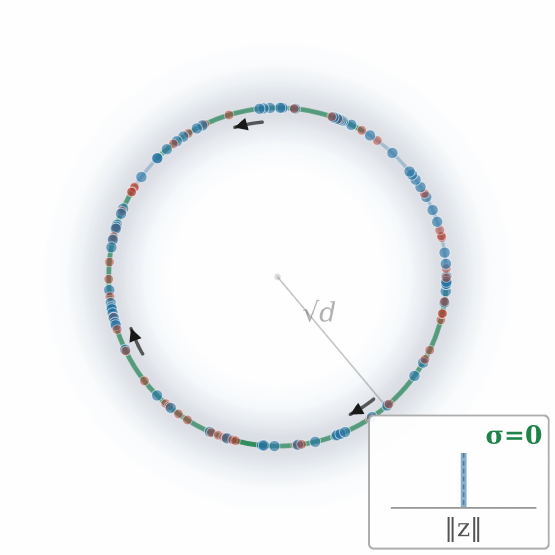}
		\caption{Spherical (Ours)}
		\label{fig:spherical-slerp}
	\end{subfigure}
	\hfill
	\begin{subfigure}[t]{0.32\textwidth}
		\centering
		\includegraphics[width=\linewidth]{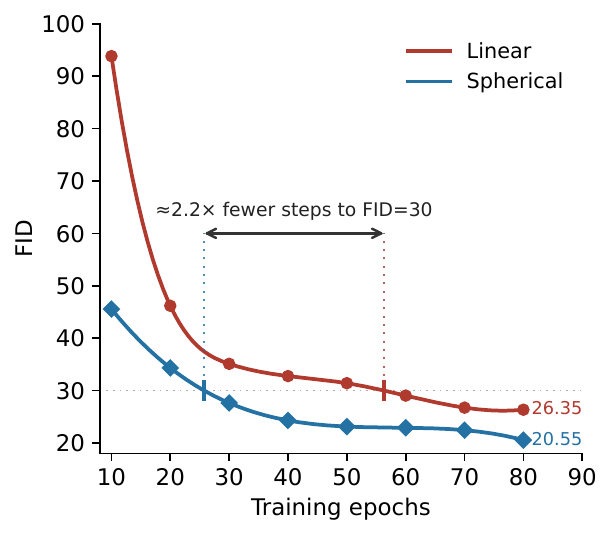}
		\caption{FID, SiT-B/2, FLUX.2 VAE}
		\label{fig:teaser-curve}
	\end{subfigure}

	\caption{\textbf{Latent flow matching ignores the geometry of VAE latents.} (a) Linear latent flow matching connects Gaussian noise to a VAE latent with a straight line. Although both endpoints concentrate on thin shells, the line passes through interior radii rarely occupied by either endpoint. (b) We project data latents and sample noise on a shared fixed-radius sphere, then train along the spherical arc, or slerp. (c) By aligning the latent geometry to the sphere, our method reaches the vanilla-linear FID$=30$ in about $2.2\times$ fewer training steps and continues to improve, without changing the diffusion architecture or adding any auxiliary encoder.}
	\label{fig:latent-geometry}
\end{figure*}

\begin{abstract}
	Latent flow matching for image generation usually transports Gaussian noise to variational autoencoder latents along linear paths. Both endpoints, however, concentrate in thin spherical shells, and a Euclidean chord leaves those shells even when preprocessing aligns their radii. By decomposing each latent token into radial and angular components, we show through component-swap probes that decoded perceptual and semantic content is carried predominantly by direction, with radius contributing much less. We therefore project data latents onto a fixed token radius, use the radial projection of Gaussian noise as the spherical prior, finetune the decoder with the encoder frozen, and replace linear interpolation with spherical linear interpolation. The resulting geodesic paths stay on the sphere at every timestep, and their velocity targets are purely angular by construction. Under matched training, the method consistently improves class-conditional ImageNet-256 FID across different image tokenizers, leaves the diffusion architecture unchanged, and requires no auxiliary encoder or representation-alignment objective.
\end{abstract}

\begin{center}
	Project Website: \url{https://aligning-latent-geometry.github.io}
\end{center}

\section{Introduction}

Diffusion~\citep{ho2020denoising, song2020score} and flow matching~\citep{lipman2023flow, liu2022flow, albergo2022building} have driven recent advances in high-fidelity image generation, predominantly through latent diffusion models that train the generator on the encodings of a pretrained variational autoencoder (VAE)~\citep{rombach2022high, esser2024scaling}. These latent image generators do not model pixels directly. They model the coordinate system produced by the VAE~\citep{labs2025FLUX2}; consequently, the shape of this latent distribution defines the space that the generator must learn to traverse~\citep{xu2026making, yao2025reconstruction}. Standard latent flow matching nevertheless treats this space as Euclidean and transports Gaussian noise to encoded data along straight lines~\citep{lipman2023flow, liu2022flow, albergo2022building, ma2024sit}. These assumptions are convenient, but they overlook two structural facts about VAE latents: their tokens concentrate in thin spherical shells, and the decoder reacts mainly to a token's direction rather than its length. We show that flow matching pays for both in image quality.

In high dimensions, both the Gaussian noise prior and VAE latents concentrate in thin spherical shells~\citep{vershynin2018high}. A straight line between two such points cuts through the interior, passing through distances from the origin that neither endpoint distribution actually occupies (\cref{sec:geometry-problem}, \cref{fig:vanilla-linear}). Within these shells, the decoder's output depends mainly on a token's direction, not its length: replacing a token's direction with that of a same-class neighbor changes the decoded image about as much as replacing the whole token, while replacing only its length barely changes it (\cref{fig:component-ablation}).

Standard linear flow matching ignores both this shell geometry and the dominance of direction over length. The velocity the model is trained to predict decomposes into a radial part that changes a token's distance from the origin and an angular part that changes its direction; near the endpoints, the radial part accounts for roughly half or more of the total (\cref{fig:flow-radial-energy}). The cost is measurable: at SiT-B/2 on FLUX.2, vanilla-linear flow matching trains more slowly and attains worse FID~\citep{heusel2017gans} after the same training budget than its spherical counterpart (\cref{fig:FLUX2-b2-ablation}, \cref{tab:transport}).

Existing geometry-aware alternatives do not supply a spherical latent for an existing pretrained VAE. Riemannian flow matching~\citep{chen2023flow} in the feature space of a frozen DINOv2 encoder~\citep{kumar2026learning} obtains spherical structure but requires an auxiliary encoder at every training and inference step. Hyperspherical methods learn a sphere-constrained encoder from scratch~\citep{davidson2018hyperspherical, xu2018spherical, yue2026image} or apply a projection only at autoregressive inference~\citep{ke2025hyperspherical}. Representation-alignment methods add losses to the VAE~\citep{yao2025reconstruction}, to diffusion training~\citep{yu2024representation}, or to both~\citep{leng2025repa}; they reshape the latent distribution rather than the geometry of the flow path through it, and require an auxiliary encoder during training.

We project each latent token onto a sphere at the encoder output of an existing pretrained VAE, finetune only the decoder for a few epochs, and replace linear interpolation with spherical linear interpolation, or slerp, between projected endpoints (\cref{fig:latent-geometry}). Because both endpoints of each token lie on the same fixed-radius sphere, the slerp arc between them stays on the sphere at every timestep, so the training target only changes the token's direction, never its length. The diffusion architecture is unchanged, no auxiliary encoder is needed at training or inference, and the projection works with both representation-aligned~\citep{yao2025reconstruction,leng2025repa} and non-aligned~\citep{labs2025FLUX2} tokenizers.

On ImageNet-256, the spherical-slerp method improves FID across FLUX.2, VA-VAE, and REPA-E FLUX.1 tokenizers under matched guidance, and the gain carries from SiT-B to SiT-XL backbones (\cref{tab:cross-vae}). A decoder-swap control rules out decoder finetuning as the explanation: swapping decoders between the vanilla and spherical pipelines degrades FID in both directions (\cref{tab:control}), showing the learned flow is tied to its latent geometry.

Our contributions are: (i) we identify a radial-shell mismatch in latent flow matching and quantify how much of its training target is spent on radial motion; (ii) we introduce a token-wise spherical projection for pretrained VAE latents with decoder-only finetuning; (iii) we train flow matching along slerp paths, with velocity targets tangent to the sphere and integration that keeps samples on it; and (iv) we show consistent ImageNet-256 gains across three tokenizer families and two model scales under matched protocols.

\section{Related Work}

\noindent \textbf{Hyperspherical latent spaces.} $L_2$-normalizing embeddings onto a fixed-radius hypersphere is a standard structural choice in discriminative representation learning~\citep{wang2017normface, deng2019arcface}. On the generative side, \citet{davidson2018hyperspherical} introduce a sphere-constrained VAE latent by replacing the Gaussian prior and posterior with the von Mises-Fisher distribution, and \citet{xu2018spherical} apply the same distribution to mitigate posterior collapse in text VAEs. For continuous-token image generation, \citet{ke2025hyperspherical} apply a fixed-radius projection to the VAE latent to stabilize variance under classifier-free guidance (CFG) in an autoregressive decoder. In concurrent work, \citet{yue2026image} train an encoder that maps images uniformly onto a sphere and generate by decoding random sphere points, bypassing diffusion entirely. These methods either modify the VAE training objective~\citep{davidson2018hyperspherical, xu2018spherical} or skip flow matching as the generator~\citep{ke2025hyperspherical, yue2026image}; none studies flow matching on the induced sphere, which is the setting we take up.

\noindent \textbf{Riemannian and manifold flow matching.} Generative modeling on manifolds was first approached through continuous normalizing flows~\citep{chen2018neural, mathieu2020riemannian} and score-based diffusion~\citep{de2022riemannian, huang2022riemannian}, which replace Euclidean drift and Brownian noise with their Riemannian counterparts and integrate along geodesics. Riemannian flow matching~\citep{chen2023flow} extends the simulation-free flow matching framework~\citep{lipman2023flow, liu2022flow, albergo2022building, rozen2021moser} to this setting, specifying conditional vector fields along geodesic interpolants and projecting velocities onto the tangent space. \citet{davis2024fisher} reparameterize categorical distributions onto the positive orthant of a sphere and train with closed-form slerp geodesics, the clearest precedent for slerp as a training-time path rather than a sampling-time interpolator. \citet{zaghen2025riemannian} introduce a curvature-dependent Jacobi-field penalty for Riemannian flow matching, and \citet{kumar2026learning} apply reweighting on the sphere induced by the final LayerNorm of a frozen DINOv2 encoder. Riemannian flow matching has seen limited use on image generation because it requires a target latent that already lies on a manifold; our spherical projection supplies such a latent space from a standard pretrained VAE without retraining the encoder.

\noindent \textbf{Representation-space diffusion.} Another line trains the generator directly in the feature space of a frozen representation encoder such as DINOv2~\citep{oquab2023dinov2}. \citet{kumar2026learning} train flow matching on the sphere induced by the LayerNorm in such an encoder, and \citet{zheng2025diffusion} pair a frozen DINO, SigLIP, or masked autoencoder encoder with a trained decoder. Another variant keeps the VAE but adds an alignment loss that pulls the generator's intermediate features toward a frozen representation encoder, either during VAE training~\citep{yao2025reconstruction}, during diffusion training~\citep{yu2024representation}, or jointly~\citep{leng2025repa}. All of these options add a training-time dependency on an auxiliary encoder, and the frozen-feature-space variants require running that encoder at inference as well. Our spherical projection imposes this sphere structure through a geometric constraint on the VAE latent, composes with both representation-aligned~\citep{yao2025reconstruction} and non-aligned~\citep{rombach2022high, labs2025FLUX2} tokenizers, and adds no auxiliary encoder at inference.

\noindent \textbf{Latent-space structure versus reconstruction fidelity.} \citet{xu2026making} and \citet{yao2025reconstruction} observe that tokenizer reconstruction quality~\citep{esser2021taming, rombach2022high} is a weak predictor of downstream diffusion generation quality, and that what governs trainability is the structure of the latent distribution. \citet{xu2026making} quantify this decoupling across published autoencoders; \citet{qiu2025robust, qiu2025image} identify sampling-error robustness as the relevant axis; \citet{yao2025reconstruction} call the phenomenon the reconstruction-generation optimization dilemma. Structural interventions proposed so far include spectral shaping of the latent~\citep{skorokhodov2025improving}, equivariance regularization~\citep{kouzelis2025eq}, end-to-end joint training~\citep{leng2025repa}, semantic regularization at scale~\citep{xiong2025gigatok}, and non-variational tokenizers with discriminative latents~\citep{chen2025masked, li2024autoregressive}. Our spherical projection is a geometric intervention in the same family: it constrains the support of the latent space rather than reshaping its spectrum or aligning it to an external target.

\section{Methodology}

\subsection{Flow Matching in Latent Space}
\label{sec:prelim}

We adopt linear-path latent flow matching as our baseline; the rest of this section examines its geometric assumptions. A pretrained autoencoder maps an image $x$ to a latent $z_1 = \mathcal{E}(x) \in \mathbb{R}^{d \times h \times w}$ with one token in $\mathbb{R}^d$ per spatial position, and decoder $\mathcal{D}$ inverts the mapping. Flow matching~\citep{lipman2023flow,liu2022flow,albergo2022building} learns a velocity field $v_\theta(z_t, t, y)$ that transports a Gaussian prior $z_0 \sim \mathcal{N}(0, I)$ to data along the linear interpolation
\begin{equation}
	\label{eq:fm-path}
	z_t = (1 - t)\, z_0 + t\, z_1, \quad t \in [0, 1],
\end{equation}
with conditional velocity $u_t = z_1 - z_0$ and objective
\begin{equation}
	\label{eq:fm-loss}
	\mathcal{L}_{\mathrm{FM}} = \mathbb{E}_{t,\, z_0,\, z_1} \left[\| v_\theta(z_t, t, y) - u_t \|^2 \right].
\end{equation}
We use the Scalable Interpolant Transformer (SiT)~\citep{ma2024sit} as the backbone. The linear interpolation in \cref{eq:fm-path} implicitly treats the latent space as $\mathbb{R}^d$ with the standard Euclidean structure, an assumption we examine in \cref{sec:geometry-problem}.

\subsection{The Geometry Problem: Concentration of Measure in High Dimensions}
\label{sec:geometry-problem}

For a standard Gaussian prior, most mass lies in a thin spherical shell near radius $\sqrt{d}$~\citep{vershynin2018high}. We define $z$ in the flow-training coordinates, after any fixed tokenizer preprocessing such as scale, shift, packing, or channel standardization. The two endpoints of the flow-matching path of \cref{eq:fm-path} are, in these coordinates, the noise sample $z_0 \sim \mathcal{N}(0, I)$ at $t=0$ and an encoded data latent $z_1$ at $t=1$. This distinction matters because raw encoder coordinates need not match the coordinates in which the noise prior and flow path are defined.

Formally, this follows from concentration of measure: for $z_0 \sim \mathcal{N}(0, I_d)$,
\begin{equation}
	\mathbb{P}\!\left(\big|\,\|z_0\| - \sqrt{d}\,\big| > t\right) \leq 2\exp(-c\,t^2),
\end{equation}
where $c > 0$ is an absolute constant \citep[Theorem~3.1.1]{vershynin2018high}. The shell width is $\mathcal{O}(1)$ regardless of $d$, so the relative thickness $\mathcal{O}(1/\sqrt{d})$ vanishes as the dimension grows. 
Although the concentration bound is conventionally centered at the RMS radius $\sqrt{d}$, the mean Gaussian radius is slightly smaller: \(\mathbb{E}\|z_0\|_2 = \sqrt{2}\Gamma((d+1)/2)/\Gamma(d/2) \approx \sqrt{d-1/2}\). We use this mean-radius expression for the analytical Gaussian rows in \cref{tab:shell-stats}; see \cref{app:gaussian-radius} for the derivation.

\begin{table}[t]
	\caption{Per-token norm statistics. Raw $\bar{r}$ is the encoder output; processed $\bar{r}$ applies each family's flow-training preprocessing. Gaussian rows use the analytical mean radius \(\mathbb{E}\|z\|_2\) and coefficient of variation for \(z\sim\mathcal{N}(0,I_d)\); see \cref{app:gaussian-radius}. VAE rows are measured on 2048 ImageNet-256 images.}
	\label{tab:shell-stats}
	\centering
	\small
	\setlength{\tabcolsep}{4pt}
		\begin{tabular}{@{}lcccccc@{}}
				\toprule
				Source        & $d$ & Preproc.       & Raw $\bar{r}$ & Raw CV & Proc. $\bar{r}$ & Proc. CV \\
			\midrule
			Gaussian Noise     & 16  & --             & 3.94          & 0.18   & 3.94            & 0.18 \\
			Gaussian Noise      & 32  & --             & 5.61          & 0.13   & 5.61            & 0.13 \\
			\midrule
			FLUX.2 VAE        & 32  & pack+std.      & 9.43          & 0.19   & 5.33            & 0.20 \\
			REPA-E FLUX.1 VAE & 16  & affine/stat.   & 9.13          & 0.22   & 1.38            & 0.23 \\
			VA-VAE        & 32  & channel std.   & 19.85         & 0.16   & 5.38            & 0.16 \\
			\bottomrule
	\end{tabular}
\end{table}

\begin{figure*}[b]
	\centering
	\begin{subfigure}[t]{0.48\textwidth}
		\centering
		\includegraphics[width=\linewidth]{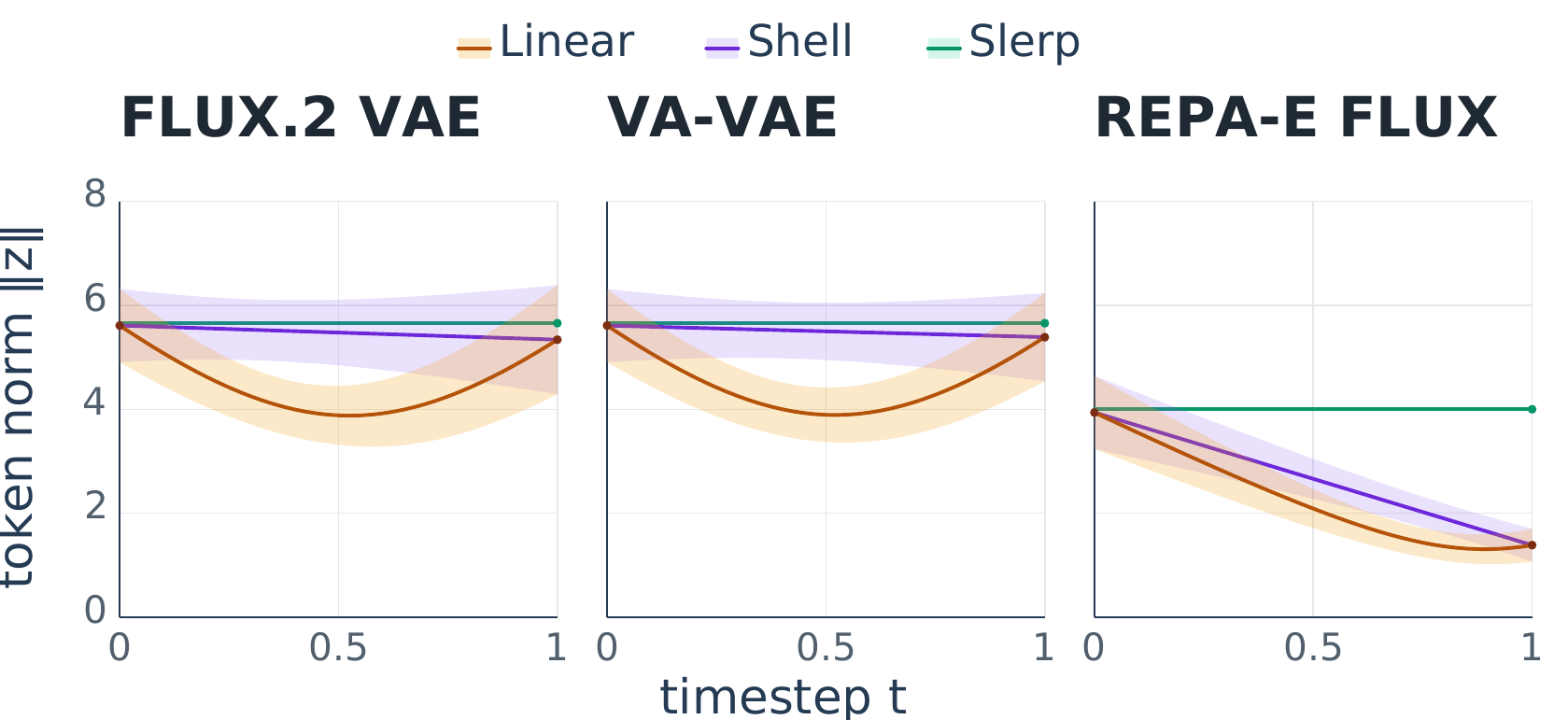}
		\caption{Per-token norm along linear, shell, and slerp paths for representative tokenizers. Lines average 2048 pairs; bands show $\pm 1\sigma$; the horizontal reference is the fixed spherical radius.}
		\label{fig:latent-norm}
	\end{subfigure}
	\hfill
	\begin{subfigure}[t]{0.48\textwidth}
		\centering
		\includegraphics[width=\linewidth]{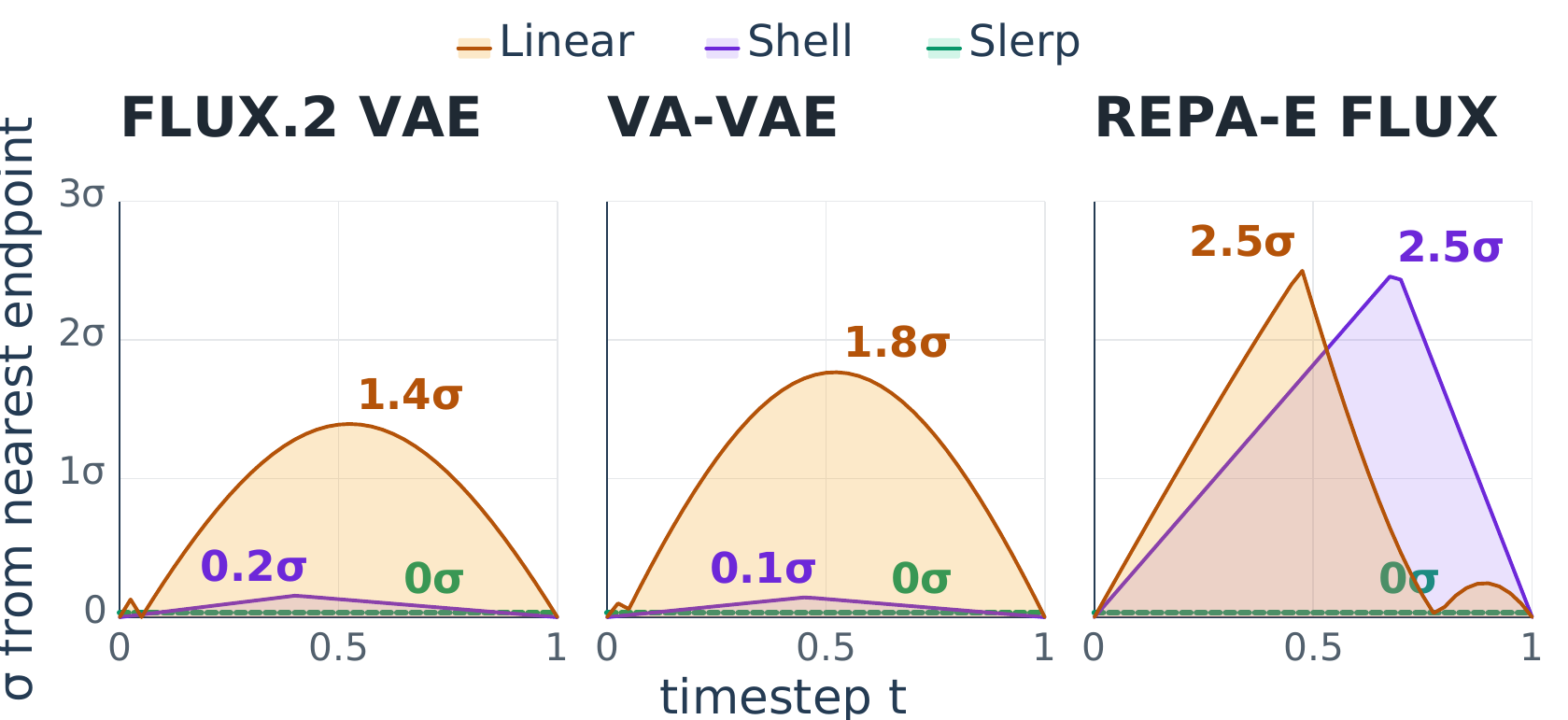}
		\caption{Off-shell distance for each path, measured in standard deviations from the nearest endpoint shell. Larger values indicate latent regions rarely occupied by either endpoint; slerp remains at the fixed spherical radius.}
		\label{fig:latent-ood}
	\end{subfigure}
	\caption{Linear paths can dip away from the endpoint shells; shell paths interpolate token radii; spherical-slerp stays on the fixed-radius sphere at every timestep.}
	\label{fig:latent-norm-ood}
\end{figure*}

The data endpoint has the same kind of radial concentration. In practice, the Kullback-Leibler (KL) term in the VAE objective~\citep{kingma2013auto} does not enforce this. Downstream latent pipelines also apply fixed preprocessing before flow training. We therefore report tokenizer norms both raw and after the preprocessing used by the vanilla baseline (\cref{fig:latent-norm}, \cref{tab:shell-stats}). Per-token norms concentrate tightly across all three tokenizers (CV $\leq 0.23$), and concentration is preserved through preprocessing (raw and processed CV agree within $0.02$). After preprocessing, FLUX.2 and VA-VAE sit close to the Gaussian shell ($\bar{r}/\sqrt{d} = 0.94$ and $0.95$), while REPA-E FLUX.1 sits well below ($\bar{r}/\sqrt{d} = 0.35$). Spherical projection collapses all three to $\bar{r}/\sqrt{d} = 1$ with CV $= 0$: preprocessing is partial, projection is universal.

Even when preprocessing brings the endpoint radii closer, a Euclidean chord through a shell still moves radially. For $z_0 = r_0 \hat{z}_0$ and $z_1 = r_1 \hat{z}_1$,
\begin{equation}
	\| (1-t)z_0 + t z_1 \|^2 = (1-t)^2 r_0^2 + t^2 r_1^2 + 2t(1-t) r_0 r_1 \langle \hat{z}_0, \hat{z}_1 \rangle.
	\label{eq:chord}
\end{equation}
Independent directions in these token dimensions have small expected cosine, so for $r_0 \approx r_1 = R$ the midpoint norm is close to $R/\sqrt{2}$ on average, and empirically the midpoint moves substantially inside the endpoint shell. If $r_0$ and $r_1$ differ, the same path also sweeps between the two shells (\cref{fig:latent-norm}).

Linear paths deviate up to $1.4\sigma$ (FLUX.2), $1.8\sigma$ (VA-VAE), and $2.5\sigma$ (REPA-E FLUX.1) from the nearest endpoint (\cref{fig:latent-norm-ood}), placing supervision on latents the training distribution rarely produces. Slerp keeps $\|z_t\| = \sqrt{d}$ throughout the flow.

To test the decoder's sensitivity to direction versus radius, we swap one component between same-class latents. For an anchor token $z_a = r_a \hat{z}_a$ and the same-position token $z_n = r_n \hat{z}_n$ from a same-class neighbor, we form $r_n \hat{z}_a$ (anchor direction, neighbor radius) and $r_a \hat{z}_n$ (anchor radius, neighbor direction), then decode. Both hybrids use real same-class components.

\begin{figure}[t]
	\centering
	\includegraphics[width=0.85\linewidth]{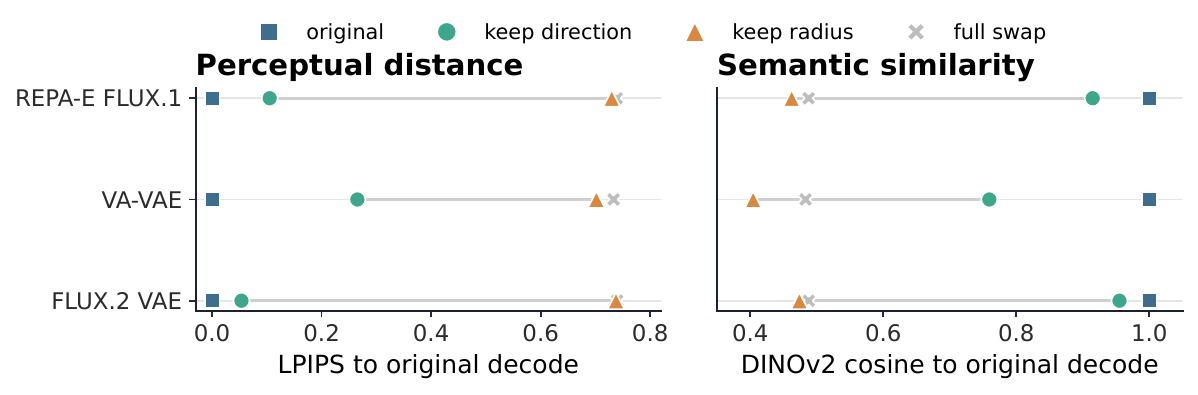}
	\caption{Angular/radial decoder sensitivity. Swapping radius with a same-class neighbor preserves the original decode, while swapping direction moves toward the neighbor; markers average 1024 anchors and shape encodes the ablation condition.}
	\label{fig:component-ablation}
\end{figure}

Keeping the anchor direction (radius swapped to the neighbor) leaves the decoded image close to the anchor, whereas keeping the anchor radius (direction swapped) moves it almost as far as replacing the whole latent with the neighbor, an asymmetry visible on both LPIPS~\citep{zhang2018unreasonable} and DINOv2 distances (\cref{fig:component-ablation}). Thus the decoder is much more sensitive to direction than to radius.

\begin{figure}[b]
	\centering
	\includegraphics[width=0.9\linewidth]{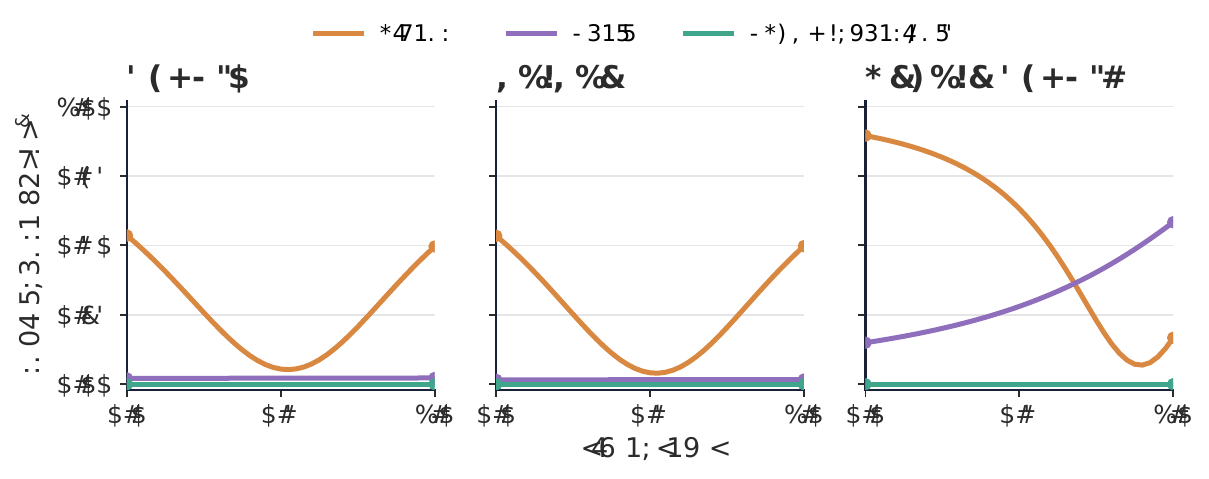}
        \vspace{-10pt}
	\caption{Radial share of the flow-matching velocity target, computed in each tokenizer's flow-training coordinates. Linear paths spend substantial supervision on radial motion, where the decoder is less sensitive; slerp on the sphere has zero radial velocity by construction.}
	\label{fig:flow-radial-energy}
\end{figure}

Linear flow matching allocates substantial supervision to radial motion, a component to which the decoder is less sensitive. Decomposing the per-token velocity target $u_t = \dot{z}_t$ into radial and tangential components in each tokenizer's flow-training coordinates yields an endpoint-dependent radial share. It is about $50\%$ at both endpoints for FLUX.2 and VA-VAE, and reaches about $90\%$ at the noise endpoint for REPA-E FLUX.1, whose data shell radius falls farthest below $\sqrt{d}$ (\cref{tab:shell-stats}, \cref{fig:flow-radial-energy}). Slerp on the sphere makes it identically zero by construction. The observed performance gap is consistent with this cost: under the matched protocol, vanilla-linear flow matching trains more slowly than spherical-slerp and attains a worse FID after the same training budget (\cref{tab:transport}).

This mismatch can be addressed either at the path level or the latent-support level. A path-level decomposition can avoid the inward chord dip by separating angular and radial motion, but it still keeps radius as a supervised prediction target. Since our component-swap probes indicate that decoded content is much more sensitive to direction than to radius, this radial target may require additional normalization or scheduling to avoid competing with angular supervision. We instead remove the radial degree of freedom at its source: project encoder outputs onto a fixed-radius sphere so that both endpoints, and the slerp geodesic between them, lie on the same sphere.

\subsection{Spherical Latent Spaces}
\label{sec:spherical}

We constrain the VAE latent space to a fixed-radius hypersphere by inserting a token-wise $L_2$ projection between the encoder and decoder. Given a pretrained encoder $\mathcal{E}$ with latent dimension $d$ and spatial resolution $h \times w$, define $\pi:\mathbb{R}^d \to \mathcal{S}^{d-1}(\sqrt{d})$ by $\pi(z) = \sqrt{d}\, z/\|z\|$ and apply it independently at each spatial position:
\begin{equation}
	z_{i,j} \leftarrow \pi(z_{i,j}) = \sqrt{d} \cdot \frac{z_{i,j}}{\|z_{i,j}\|}, \quad (i,j) \in \{1,\ldots,h\} \times \{1,\ldots,w\}.
\end{equation}
Each token then satisfies $\|z_{i,j}\| = \sqrt{d}$. The radius $\sqrt{d}$ matches the concentration radius of a standard Gaussian in $d$ dimensions, aligning the projected latent scale with the noise prior; the full latent tensor lives on a product of $h \cdot w$ copies of $\mathcal{S}^{d-1}(\sqrt{d})$. This setting differs from prior hyperspherical VAE work in two respects: we constrain existing pretrained Gaussian VAEs with a hard projection rather than training an encoder from scratch, and we keep the downstream flow matching model rather than replacing it with direct decoding from the sphere.

We freeze the encoder, insert the projection at the encoder output, and finetune only the decoder and discriminator for the tokenizers used in the generation experiments: FLUX.2 \citep{labs2025FLUX2} ($f{=}8$), VA-VAE \citep{yao2025reconstruction} ($f{=}16$, DINOv2-aligned), and REPA-E FLUX.1 \citep{leng2025repa} ($f{=}8$). The reconstruction objective retains the original pixel $L_1$, LPIPS, and patch-level adversarial losses~\citep{isola2017image, esser2021taming}; the KL term is dropped, since the encoder is frozen and the projected latent is deterministic. For VA-VAE the DINOv2 alignment terms have zero gradient once the encoder is frozen, so they are removed; the alignment learned during pretraining persists in the frozen encoder weights. We finetune for five epochs on ImageNet and report the reconstruction tradeoff in \cref{tab:rfid}.

Tokenizer reconstruction quality is a weak predictor of downstream diffusion FID across published autoencoders \citep{xu2026making, qiu2025robust, yao2025reconstruction}; \citet{skorokhodov2025improving} trace this to structural properties of the latent rather than decoder fidelity. The radial-shell gap is a structural property of the same kind: a feature of the latent that affects flow matching while leaving reconstruction quality close to the finetuned vanilla control. We therefore report rFID and FID separately and quantify the tradeoff in \cref{sec:experiments}. The component ablation (\cref{fig:component-ablation}, with population-mean substitute and per-sample distributions in \cref{fig:component-ablation-global,fig:perturb,fig:perturb-sameclass}) shows the decoder reads direction far more strongly than radius, so fixing the radius discards the component the decoder is less sensitive to.

The projection is applied only inside the VAE: the diffusion model (SiT) sees the spherical latents as ordinary vectors in $\mathbb{R}^d$ and requires no architectural changes; no auxiliary encoder is used during diffusion training or inference. This distinguishes the construction from methods that achieve spherical latent geometry by diffusing in the feature space of a frozen DINOv2 \citep{kumar2026learning, oquab2023dinov2}, which run the auxiliary encoder on every generated sample. With both endpoints fixed on $\mathcal{S}^{d-1}(\sqrt{d})$, the remaining design choice is the transport between them.

\subsection{Transport on the Sphere}
\label{sec:transport}

With both endpoints on $\mathcal{S}^{d-1}(\sqrt{d})$, slerp gives the shortest geodesic between them and stays on the sphere for all $t$. We compare it with linear and shell paths, which leave the fixed-radius sphere, to separate the effect of geodesic transport from the projection itself. The standard Gaussian prior concentrates near radius $\sqrt{d}$, but its samples do not lie exactly on the sphere. For the spherical path, we remove only this radial fluctuation: for each token we sample $\epsilon \sim \mathcal{N}(0, I_d)$ and set
\begin{equation}
	z_0 = \sqrt{d}\,\frac{\epsilon}{\|\epsilon\|_2}.
\end{equation}
By rotational invariance of the isotropic Gaussian, $\epsilon/\|\epsilon\|_2$ is uniform on $\mathcal{S}^{d-1}$, so $z_0 \sim \mathrm{Uniform}(\mathcal{S}^{d-1}(\sqrt{d}))$; see \cref{app:projected-gaussian-uniform}. The spherical noise endpoint is thus the angular component of the same Gaussian prior used by the Euclidean baseline. The resulting uniform spherical distribution is also the standard prior used in Riemannian flow matching and directional statistics \citep{chen2023flow, mardia2009directional}.

\paragraph{Linear path (baseline).} The Euclidean interpolation $z_t = (1-t)\,z_0 + t\,z_1$ from \cref{sec:prelim} applies to spherical latents without modification: both endpoints are on the sphere but the path leaves the sphere at intermediate times. This baseline isolates the effect of the spherical constraint from the effect of geometry-aware transport.

\begin{wrapfigure}[16]{r}{0.32\textwidth}
	\centering
	\vspace{-1.0em}
	\includegraphics[width=\linewidth]{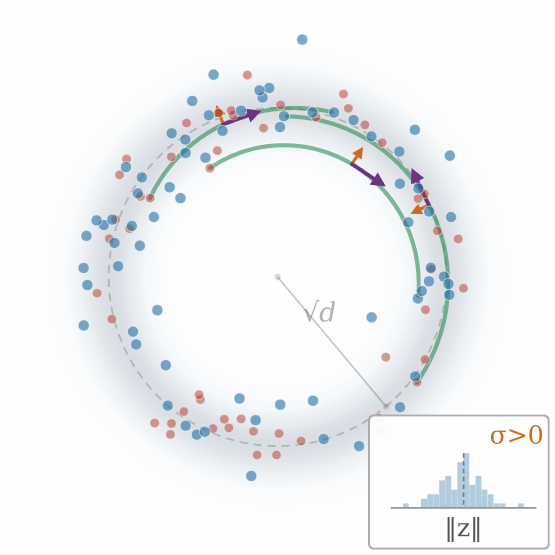}
	\caption{Shell-decomposed path: radius and direction interpolated separately.}
	\label{fig:shell-flow}
\end{wrapfigure}

\paragraph{Shell path.} Without requiring the spherical VAE constraint, each endpoint is decomposed into a direction and a magnitude, $z = r\,\hat{z}$ with $\hat{z} = z/\|z\|$ and $r = \|z\|$, and the two are interpolated separately (\cref{fig:shell-flow}):
\begin{equation}
	\hat{z}_t = \mathrm{slerp}(\hat{z}_0, \hat{z}_1;\, t), \qquad r_t = (1-t)\,r_0 + t\,r_1, \qquad z_t = r_t\,\hat{z}_t.
\end{equation}
Because $r_t$ moves linearly from $r_0$ to $r_1$, the path avoids the inward chord dip of Euclidean interpolation (\cref{fig:latent-norm}). However, it still preserves radius as a supervised component of the flow target. The model is asked to learn both angular motion, which our component-swap probes indicate is more relevant to decoded content (\cref{fig:component-ablation}), and radial motion, which is less decoder-sensitive. Without a separate normalization or schedule, the joint objective may allocate substantial capacity to the radial component even though it contributes less to the decoded image. Spherical projection takes the complementary approach: rather than balancing angular and radial targets, we remove the radial degree of freedom before flow training, so the resulting slerp target is purely angular.

\paragraph{Slerp path.} When both endpoints lie on $\mathcal{S}^{d-1}(\sqrt{d})$, we interpolate along the spherical linear interpolation geodesic~\citep{shoemake1985animating}. Writing $\hat{z} = z/\|z\|$ so that $z = \sqrt{d}\,\hat{z}$, and given angular separation $\omega = \arccos\langle \hat{z}_0, \hat{z}_1 \rangle$, the geodesic is
\begin{equation}
	z_t = \sqrt{d} \cdot \mathrm{slerp}(\hat{z}_0, \hat{z}_1;\, t) = \sqrt{d} \left(\frac{\sin((1-t)\,\omega)}{\sin\omega}\,\hat{z}_0 + \frac{\sin(t\,\omega)}{\sin\omega}\,\hat{z}_1\right),
\end{equation}
the unique shortest path between $z_0$ and $z_1$ on the sphere for $\omega < \pi$~\citep{do1992riemannian}, staying on the sphere at all times. Following \citet{chen2023flow}, the conditional velocity field is the time derivative of the geodesic projected onto the tangent space $T_{z_t}\mathcal{S}^{d-1}(\sqrt{d})$:
\begin{equation}
	u_t = \frac{d}{dt}\,z_t, \qquad u_t \leftarrow u_t - \frac{\langle u_t, z_t\rangle}{\|z_t\|^2}\,z_t.
\end{equation}
In exact arithmetic the slerp time derivative already lies in $T_{z_t}\mathcal{S}^{d-1}(\sqrt{d})$, so the projection on the target acts as a numerical safeguard against finite-precision drift. Its more substantive role is on the model output: the network $v_\theta$ has no architectural tangency constraint, so the same projection is applied before the squared-error loss to enforce that the learned field is tangent and to guarantee sphere-preserving sampling \citep{chen2023flow, rozen2021moser}. The resulting target is purely angular (\cref{fig:flow-radial-energy}).

For the slerp path, the training loss is the flow-matching objective with both the target and model velocity projected to the tangent space:
\begin{equation}
	\mathcal{L}_{\mathrm{slerp}} = \mathbb{E}_{t,\,z_0,\,z_1}\!\left[\left\|\Pi_{z_t}\,v_\theta(z_t, t, y) - \Pi_{z_t}\,u_t\right\|^2\right], \qquad \Pi_{z_t}\,v = v - \frac{\langle v, z_t\rangle}{\|z_t\|^2}\,z_t.
\end{equation}
At inference we apply the same tangent projection to the model velocity and integrate with the exponential map: $z_{t+\Delta t} = \exp_{z_t}\!\left(\Pi_{z_t}\,v_\theta(z_t, t, y) \cdot \Delta t\right)$ stays on the sphere by construction and has a closed-form trigonometric expression, adding negligible cost. This sampler is matched to the slerp training target. Along a slerp path with endpoint angle $\omega$, the target velocity is a constant-speed tangent vector of norm $\sqrt{d}\,\omega$. A single exponential-map step with a perfect predictor moves by geodesic arc length $\Delta t\,\sqrt{d}\,\omega$ and therefore lands exactly on the same slerp curve at time $t+\Delta t$; an Euler step followed by radial projection stays on the same great circle but advances only by arc length $\sqrt{d}\,\arctan(\Delta t\,\omega)$, producing a one-step deficit $\sqrt{d}\,\bigl(\Delta t\,\omega - \arctan(\Delta t\,\omega)\bigr)$. \Cref{sec:experiments} evaluates the resulting latent-space and transport choices.

\section{Experiments}
\label{sec:experiments}

We evaluate the spherical-slerp method against the vanilla-linear baseline under matched training, asking whether the latent-geometry gain holds across tokenizer families and backbone scales.

\noindent \textbf{Experimental Setup.}\label{sec:setup} We hold the architecture, data, training budget, and evaluator fixed so that observed differences are attributable to the latent geometry and transport path. We use ImageNet-256 class-conditional generation \citep{russakovsky2015imagenet} with the SiT family \citep{ma2024sit}: B/2 and XL/2 on FLUX.2, B/1 and XL/1 on VA-VAE, and B/2 on REPA-E FLUX.1, adjusting patch size with each tokenizer's downsampling factor so that all settings yield 256 tokens per $256 \times 256$ image. Models train from scratch for $80$ epochs at batch size $256$ in bfloat16 with AdamW \citep{loshchilov2017decoupled}. Latents are precomputed once per VAE after the tokenizer-specific preprocessing used by the vanilla flow baseline; the spherical variant applies the token-wise projection in those same coordinates before decoder finetuning and flow training. The encoder, and spherical projection are frozen throughout diffusion training. Key ablations are in the supplementary. All reported generation FID numbers are FID-50K, computed on 50K generated samples against the ImageNet-256 training reference. Final generation results use 250 sampling steps; training-curve and latent-path ablations use the same evaluator with 50 sampling steps.

\begin{wraptable}[11]{r}{0.67\textwidth}
		\caption{Latent-support and transport-path ablation on SiT-B/2 with the FLUX.2 VAE.}
	\label{tab:transport}
	\centering
	\small
	\begin{tabular}{@{}lllllr@{}}
		\toprule
		\multicolumn{5}{c}{Training and sampling protocol} & FID$\downarrow$ \\
		\cmidrule(lr){1-5}\cmidrule(l){6-6}
		Latent    & Prior    & Path   & Tan. proj. & Sampler  & CFG$=1.0$ \\
		\midrule
		Vanilla   & Gaussian & Linear & off           & Euler    & 26.35     \\
		Vanilla   & Gaussian & Shell  & off           & Euler    & 27.26     \\
		Spherical & Uniform  & Linear & off           & Euler    & 22.85     \\
		\rowcolor{gray!20} Spherical & Uniform  & Slerp  & on            & Exp. map & 20.55     \\
		\bottomrule
	\end{tabular}
\end{wraptable}

\noindent \textbf{Reconstruction cost of spherical VAEs.}\label{sec:rfid} Though spherical projection introduces a modest reconstruction cost, relative to the matched finetuned-vanilla (FT-vanilla) VAE, the downstream generation performance improves noticeably (\cref{tab:transport}). Downstream generation always decodes through the corresponding spherical decoder whenever spherical latents are used, so the gain cannot be attributed to using a higher-capacity decoder (\cref{tab:control}). Instead, the comparison isolates the effect of changing the latent support and transport geometry.

\begin{wraptable}[11]{r}{0.61\textwidth}
	\caption{ImageNet-256 validation rFID. Spherical uses the fixed-radius projection with decoder/discriminator finetuning; finetuned vanilla (FT-vanilla) is the no-projection control.}
	\label{tab:rfid}
	\centering
	\small
	\begin{tabular}{@{}lccccc@{}}
		\toprule
		\multicolumn{3}{c}{Tokenizer} & \multicolumn{3}{c}{rFID$\downarrow$} \\
		\cmidrule(lr){1-3}
		\cmidrule(lr){4-6}
		VAE           & $d$ & $f$ & Original & FT-vanilla & Spherical \\
		\midrule
		FLUX.2        & 32  & 8   & 0.186    & 0.486      & 1.072     \\
		VA-VAE        & 32  & 16  & 0.298    & 1.116      & 1.064     \\
		REPA-E FLUX.1 & 16  & 8   & 0.209    & 0.759      & 0.824     \\
		\bottomrule
	\end{tabular}
\end{wraptable}

\begin{wrapfigure}[18]{r}{0.6\textwidth}
	\centering
	\includegraphics[width=\linewidth]{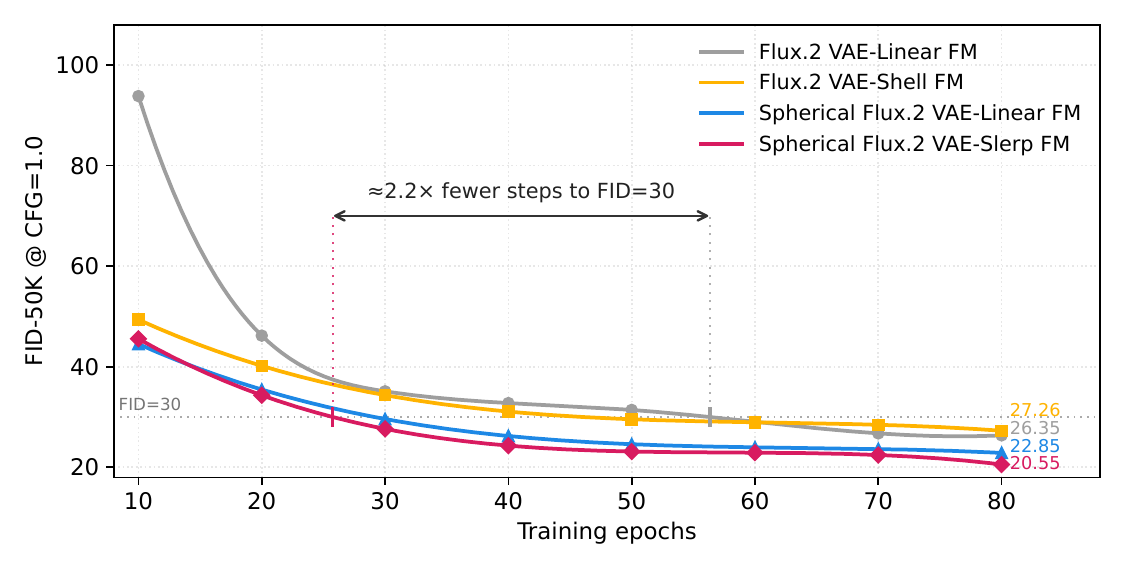}
	\caption{FID-50K at CFG$=1.0$ for the latent-support and transport-path ablation on SiT-B/2 with the FLUX.2 VAE. Spherical-slerp reaches FID$=30$ in about $2.2{\times}$ fewer training steps than vanilla-linear and continues to improve.}
	\label{fig:FLUX2-b2-ablation}
\end{wrapfigure}

\noindent \textbf{Transport path comparison.}\label{sec:transport-ablation} On SiT-B/2 with the FLUX.2 VAE, the spherical construction, comprising the fixed-radius projection, the uniform spherical prior, and the spherical-decoder finetune, explains most of the gain over vanilla-linear, and the slerp path on top gives the best observed result (\cref{tab:transport}). The Spherical-Linear row isolates this construction under a Euclidean transport; the Spherical-Slerp row adds the geometry-matched path. Shell-decomposed transport on vanilla latents, by contrast, does not improve the baseline. Avoiding the inward chord dip alone is therefore insufficient: with vanilla latents the network is still trained to predict both angular and radial velocity under a single objective. Component-swap probes show that decoded content is far more sensitive to angular direction than to radius (\cref{fig:component-ablation}), so the radial target competes with the more generation-relevant angular one; coexisting cleanly would require separate normalization, reweighting, or scheduling. Spherical-slerp removes the conflict by training only on tangent, angular velocity, and reaches the target FID in fewer training steps (\cref{fig:FLUX2-b2-ablation}).

\begin{wraptable}[7]{r}{0.45\textwidth}
	\vspace{-5pt}
	\caption{ImageNet-256 FID-50K at 200 epochs on SiT-B/2 with the FLUX.2 VAE.}
	\label{tab:long-training}
	\centering
	\small
	\begin{tabular}{@{}lcc@{}}
		\toprule
		Method                                & CFG$=1.0$ & CFG$=1.5$ \\
		\midrule
		Vanilla-linear                        & 9.15      & 3.22      \\
		\rowcolor{gray!20} Spherical-slerp    & 8.35      & 2.91      \\
		\bottomrule
	\end{tabular}
\end{wraptable}

\noindent \textbf{Generation across tokenizers and scales.}\label{sec:cross-vae} The spherical-slerp gains hold across tokenizer families, model scales, and guidance settings. It improves matched-CFG FID on FLUX.2 and VA-VAE at both B and XL scales, and on REPA-E FLUX.1 at every tested guidance setting (\cref{tab:cross-vae}). The tokenizers differ in objective, channel count, and downsampling factor, so the gain is a property of the latent geometry rather than of a particular VAE.

\begin{table}[htbp]
	\caption{ImageNet-256 FID-50K and IS across tokenizers and scales; each row compares vanilla-linear and spherical-slerp at the same CFG with 80 epochs training.}
	\label{tab:cross-vae}
	\centering
	\small
	\begin{tabular}{@{}llcr>{\columncolor{gray!20}}rrr>{\columncolor{gray!20}}r@{}}
		\toprule
		\multicolumn{3}{c}{Setting} & \multicolumn{3}{c}{FID$\downarrow$} & \multicolumn{2}{c}{IS$\uparrow$} \\
		\cmidrule(lr){1-3}\cmidrule(lr){4-6}\cmidrule(l){7-8}
		Tokenizer  & Scale & CFG & Linear & Slerp & Rel.\ $\Delta$ & Linear & Slerp \\
		\midrule
		\multirow{4}{*}{FLUX.2}        & \multirow{2}{*}{B/2}  & 1.0 & 26.14 & 20.21 & $-22.7\%$ & 52.5  & 66.5  \\
		                               &                       & 1.5 & 9.99  & 6.29  & $-37.0\%$ & 141.8 & 169.7 \\
		\cmidrule(lr){2-8}
		                               & \multirow{2}{*}{XL/2} & 1.0 & 13.00 & 10.71 & $-17.6\%$ & 85.4  & 102.3 \\
		                               &                       & 1.5 & 4.32  & 3.95  & $-8.6\%$  & 208.1 & 238.9 \\
		\midrule
		\multirow{5}{*}{VA-VAE}        & \multirow{3}{*}{B/1}  & 1.0 & 26.99 & 21.96 & $-18.6\%$ & 49.0  & 58.2  \\
		                               &                       & 1.5 & 10.86 & 7.81  & $-28.1\%$ & 120.7 & 143.4 \\
		                               &                       & 2.0 & 6.84  & 5.46  & $-20.2\%$ & 199.6 & 231.6 \\
		\cmidrule(lr){2-8}
		                               & \multirow{2}{*}{XL/1} & 1.0 & 12.16 & 10.36 & $-14.8\%$ & 93.0  & 107.3 \\
		                               &                       & 1.5 & 4.23  & 3.88  & $-8.2\%$  & 227.9 & 263.2 \\
		\midrule
		\multirow{3}{*}{REPA-E FLUX.1} & \multirow{3}{*}{B/2}  & 1.0 & 38.00 & 26.07 & $-31.4\%$ & 42.9  & 63.1  \\
		                               &                       & 1.5 & 13.83 & 6.88  & $-50.2\%$ & 118.3 & 172.1 \\
		                               &                       & 2.0 & 7.58  & 5.43  & $-28.4\%$ & 204.2 & 274.0 \\
		\bottomrule
	\end{tabular}
\end{table}

\noindent \textbf{Scaling.}\label{sec:scaling} The XL-scale rows in \cref{tab:cross-vae} show that the B-scale recipe transfers to larger backbones. At CFG$=1.0$, spherical-slerp improves FLUX.2 XL/2 by $17.6\%$ and VA-VAE XL/1 by $14.8\%$; at CFG$=1.5$, it remains ahead on both tokenizer families. The gains hold without changing the diffusion architecture or adding an auxiliary encoder during diffusion training or inference.

The matched-protocol comparison in \cref{sec:cross-vae} fixes the training budget at 80 epochs to control compute across tokenizers and scales. To confirm that the spherical-slerp advantage is not specific to short-budget training, we extend FLUX.2 B/2 to 200 epochs at the same batch size and otherwise identical settings (\cref{tab:long-training}). Both methods improve substantially over their 80-epoch numbers, and spherical-slerp remains ahead at both guidance settings: $8.35$ versus $9.15$ at CFG$=1.0$ and $2.91$ versus $3.22$ at CFG$=1.5$. The lead persists at the longer training budget, consistent with the training-curve view in \cref{fig:FLUX2-b2-ablation}.

\section{Conclusion}

In this paper, we identify a geometric mismatch in latent flow matching: VAE latents and Gaussian noise concentrate on thin shells, while linear paths leave those shells and allocate supervision to radial motion to which the decoder is weakly sensitive. We address this with spherical latent flow matching: project each VAE token to a fixed-radius sphere, use projected Gaussian directions as a uniform spherical prior, and train along slerp geodesics with tangent velocities, without changing the diffusion architecture or adding an auxiliary encoder. On ImageNet-256, this geometry improves matched-CFG FID across FLUX.2, VA-VAE, and REPA-E FLUX.1, transfers from SiT-B to SiT-XL, and reaches the vanilla-linear FID floor in fewer training steps. These results establish latent support geometry as a useful design axis; future work should test spherical constraints under stronger compression and design sphere-constrained tokenizers that match the reconstruction quality of unconstrained tokenizers.

\section{Acknowledgments} Pinar Yanardag is supported National
Science Foundation under Grant No. 2543524.

\bibliographystyle{unsrtnat}
\bibliography{main}

@article{albergo2022building,
  title   = {Building normalizing flows with stochastic interpolants},
  author  = {Albergo, Michael S and Vanden-Eijnden, Eric},
  journal = {arXiv preprint arXiv:2209.15571},
  year    = {2022}
}

@article{chen2018neural,
  author  = {Chen, Ricky TQ and Rubanova, Yulia and Bettencourt, Jesse and Duvenaud, David K},
  journal = {Advances in neural information processing systems},
  title   = {Neural ordinary differential equations},
  volume  = {31},
  year    = {2018}
}

@article{chen2023flow,
  title   = {Flow matching on general geometries},
  author  = {Chen, Ricky TQ and Lipman, Yaron},
  journal = {arXiv preprint arXiv:2302.03660},
  year    = {2023}
}

@inproceedings{chen2025masked,
  title     = {Masked autoencoders are effective tokenizers for diffusion models},
  author    = {Chen, Hao and Han, Yujin and Chen, Fangyi and Li, Xiang and Wang, Yidong and Wang, Jindong and Wang, Ze and Liu, Zicheng and Zou, Difan and Raj, Bhiksha},
  booktitle = {Forty-second International Conference on Machine Learning},
  year      = {2025}
}

@article{davidson2018hyperspherical,
  title   = {Hyperspherical variational auto-encoders},
  author  = {Davidson, Tim R and Falorsi, Luca and De Cao, Nicola and Kipf, Thomas and Tomczak, Jakub M},
  journal = {arXiv preprint arXiv:1804.00891},
  year    = {2018}
}

@article{davis2024fisher,
  title   = {Fisher flow matching for generative modeling over discrete data},
  author  = {Davis, Oscar and Kessler, Samuel and Petrache, Mircea and Ceylan, {\.I}smail {\.I}lkan and Bronstein, Michael and Bose, Avishek J},
  journal = {Advances in Neural Information Processing Systems},
  volume  = {37},
  pages   = {139054--139084},
  year    = {2024}
}

@article{de2022riemannian,
  title   = {Riemannian score-based generative modelling},
  author  = {De Bortoli, Valentin and Mathieu, Emile and Hutchinson, Michael and Thornton, James and Teh, Yee Whye and Doucet, Arnaud},
  journal = {Advances in neural information processing systems},
  volume  = {35},
  pages   = {2406--2422},
  year    = {2022}
}

@inproceedings{deng2019arcface,
  title     = {Arcface: Additive angular margin loss for deep face recognition},
  author    = {Deng, Jiankang and Guo, Jia and Xue, Niannan and Zafeiriou, Stefanos},
  booktitle = {Proceedings of the IEEE/CVF conference on computer vision and pattern recognition},
  pages     = {4690--4699},
  year      = {2019}
}

@book{do1992riemannian,
  title     = {Riemannian geometry},
  author    = {Do Carmo, Manfredo Perdigao and Flaherty Francis, J},
  volume    = {393},
  year      = {1992},
  publisher = {Springer}
}

@inproceedings{esser2021taming,
  author    = {Esser, Patrick and Rombach, Robin and Ommer, Bj{\"o}rn},
  booktitle = {Proceedings of the IEEE/CVF conference on computer vision and pattern recognition},
  pages     = {12873--12883},
  title     = {Taming transformers for high-resolution image synthesis},
  year      = {2021}
}

@inproceedings{esser2024scaling,
  author    = {Esser, Patrick and Kulal, Sumith and Blattmann, Andreas and Entezari, Rahim and M{\"u}ller, Jonas and Saini, Harry and Levi, Yam and Lorenz, Dominik and Sauer, Axel and Boesel, Frederic and others},
  booktitle = {International Conference on Machine Learning},
  title     = {Scaling rectified flow transformers for high-resolution image synthesis},
  year      = {2024}
}

@article{heusel2017gans,
  title   = {{GANs} trained by a two time-scale update rule converge to a local {Nash} equilibrium},
  author  = {Heusel, Martin and Ramsauer, Hubert and Unterthiner, Thomas and Nessler, Bernhard and Hochreiter, Sepp},
  journal = {Advances in neural information processing systems},
  volume  = {30},
  year    = {2017}
}

@article{ho2020denoising,
  author  = {Ho, Jonathan and Jain, Ajay and Abbeel, Pieter},
  journal = {Advances in Neural Information Processing Systems},
  pages   = {6840--6851},
  title   = {Denoising diffusion probabilistic models},
  volume  = {33},
  year    = {2020}
}

@article{huang2022riemannian,
  title   = {Riemannian diffusion models},
  author  = {Huang, Chin-Wei and Aghajohari, Milad and Bose, Joey and Panangaden, Prakash and Courville, Aaron C},
  journal = {Advances in Neural Information Processing Systems},
  volume  = {35},
  pages   = {2750--2761},
  year    = {2022}
}

@inproceedings{isola2017image,
  author    = {Isola, Phillip and Zhu, Jun-Yan and Zhou, Tinghui and Efros, Alexei A},
  booktitle = {Proceedings of the IEEE conference on computer vision and pattern recognition},
  pages     = {1125--1134},
  title     = {Image-to-image translation with conditional adversarial networks},
  year      = {2017}
}

@article{ke2025hyperspherical,
  title   = {Hyperspherical latents improve continuous-token autoregressive generation},
  author  = {Ke, Guolin and Xue, Hui},
  journal = {arXiv preprint arXiv:2509.24335},
  year    = {2025}
}

@inproceedings{kingma2013auto,
  author    = {Diederik P. Kingma and Max Welling},
  booktitle = {ICLR},
  title     = {Auto-Encoding Variational {Bayes}.},
  url       = {http://arxiv.org/abs/1312.6114},
  year      = {2014}
}

@article{kouzelis2025eq,
  title   = {{EQ-VAE}: Equivariance regularized latent space for improved generative image modeling},
  author  = {Kouzelis, Theodoros and Kakogeorgiou, Ioannis and Gidaris, Spyros and Komodakis, Nikos},
  journal = {arXiv preprint arXiv:2502.09509},
  year    = {2025}
}

@article{kumar2026learning,
  title   = {Learning on the Manifold: Unlocking Standard Diffusion Transformers with Representation Encoders},
  author  = {Kumar, Amandeep and Patel, Vishal M},
  journal = {arXiv preprint arXiv:2602.10099},
  year    = {2026}
}

@misc{labs2025flux2,
  author       = {{Black Forest Labs}},
  title        = {{FLUX.2: Frontier Visual Intelligence}},
  year         = {2025},
  howpublished = {\url{https://bfl.ai/blog/flux-2}}
}

@inproceedings{leng2025repa,
  title     = {{REPA-E}: Unlocking {VAE} for end-to-end tuning of latent diffusion transformers},
  author    = {Leng, Xingjian and Singh, Jaskirat and Hou, Yunzhong and Xing, Zhenchang and Xie, Saining and Zheng, Liang},
  booktitle = {Proceedings of the IEEE/CVF International Conference on Computer Vision},
  pages     = {18262--18272},
  year      = {2025}
}

@article{li2024autoregressive,
  title   = {Autoregressive image generation without vector quantization},
  author  = {Li, Tianhong and Tian, Yonglong and Li, He and Deng, Mingyang and He, Kaiming},
  journal = {Advances in Neural Information Processing Systems},
  volume  = {37},
  pages   = {56424--56445},
  year    = {2024}
}

@inproceedings{lipman2023flow,
  author    = {Lipman, Yaron and Chen, Ricky TQ and Ben-Hamu, Heli and Nickel, Maximilian and Le, Matt},
  booktitle = {International Conference on Learning Representations},
  title     = {Flow matching for generative modeling},
  year      = {2023}
}

@inproceedings{liu2022flow,
  author    = {Liu, Xingchao and Gong, Chengyue and Liu, Qiang},
  booktitle = {International Conference on Learning Representations},
  title     = {Flow straight and fast: Learning to generate and transfer data with rectified flow},
  year      = {2023}
}

@inproceedings{loshchilov2017decoupled,
  author    = {Ilya Loshchilov and Frank Hutter},
  booktitle = {ICLR},
  title     = {Decoupled Weight Decay Regularization.},
  url       = {https://openreview.net/forum?id=Bkg6RiCqY7},
  year      = {2019}
}

@inproceedings{ma2024sit,
  author       = {Ma, Nanye and Goldstein, Mark and Albergo, Michael S and Boffi, Nicholas M and Vanden-Eijnden, Eric and Xie, Saining},
  booktitle    = {European Conference on Computer Vision},
  organization = {Springer},
  pages        = {23--40},
  title        = {Sit: Exploring flow and diffusion-based generative models with scalable interpolant transformers},
  year         = {2024}
}

@book{mardia2009directional,
  title     = {Directional statistics},
  author    = {Mardia, Kanti V and Jupp, Peter E},
  year      = {2009},
  publisher = {John Wiley \& Sons}
}

@article{mathieu2020riemannian,
  title   = {Riemannian continuous normalizing flows},
  author  = {Mathieu, Emile and Nickel, Maximilian},
  journal = {Advances in neural information processing systems},
  volume  = {33},
  pages   = {2503--2515},
  year    = {2020}
}

@article{oquab2023dinov2,
  author  = {M. Oquab and Timothée Darcet and Théo Moutakanni and Huy V. Vo and Marc Szafraniec and Vasil Khalidov and Pierre Fernandez and Daniel Haziza and Francisco Massa and Alaaeldin El-Nouby and Mahmoud Assran and Nicolas Ballas and Wojciech Galuba and Russ Howes and Po-Yao (Bernie) Huang and Shang-Wen Li and Ishan Misra and Michael G. Rabbat and Vasu Sharma and Gabriel Synnaeve and Hu Xu and H. Jégou and J. Mairal and Patrick Labatut and Armand Joulin and Piotr Bojanowski},
  doi     = {10.48550/arxiv.2304.07193},
  journal = {Trans. Mach. Learn. Res.},
  title   = {{DINOv2}: Learning Robust Visual Features without Supervision},
  url     = {https://doi.org/10.48550/arxiv.2304.07193},
  year    = {2023}
}

@article{qiu2025image,
  title   = {Image tokenizer needs post-training},
  author  = {Qiu, Kai and Li, Xiang and Chen, Hao and Kuen, Jason and Xu, Xiaohao and Gu, Jiuxiang and Luo, Yinyi and Raj, Bhiksha and Lin, Zhe and Savvides, Marios},
  journal = {arXiv preprint arXiv:2509.12474},
  year    = {2025}
}

@article{qiu2025robust,
  title   = {Robust latent matters: Boosting image generation with sampling error synthesis},
  author  = {Qiu, Kai and Li, Xiang and Kuen, Jason and Chen, Hao and Xu, Xiaohao and Gu, Jiuxiang and Luo, Yinyi and Raj, Bhiksha and Lin, Zhe and Savvides, Marios},
  journal = {arXiv preprint arXiv:2503.08354},
  year    = {2025}
}

@inproceedings{rombach2022high,
  author    = {Rombach, Robin and Blattmann, Andreas and Lorenz, Dominik and Esser, Patrick and Ommer, Bj{\"o}rn},
  booktitle = {Proceedings of the IEEE/CVF Conference on Computer Vision and Pattern Recognition},
  pages     = {10684--10695},
  title     = {High-resolution image synthesis with latent diffusion models},
  year      = {2022}
}

@article{rozen2021moser,
  title   = {Moser flow: Divergence-based generative modeling on manifolds},
  author  = {Rozen, Noam and Grover, Aditya and Nickel, Maximilian and Lipman, Yaron},
  journal = {Advances in neural information processing systems},
  volume  = {34},
  pages   = {17669--17680},
  year    = {2021}
}

@article{russakovsky2015imagenet,
  title     = {{ImageNet} large scale visual recognition challenge},
  author    = {Russakovsky, Olga and Deng, Jia and Su, Hao and Krause, Jonathan and Satheesh, Sanjeev and Ma, Sean and Huang, Zhiheng and Karpathy, Andrej and Khosla, Aditya and Bernstein, Michael and others},
  journal   = {International journal of computer vision},
  volume    = {115},
  number    = {3},
  pages     = {211--252},
  year      = {2015},
  publisher = {Springer}
}

@inproceedings{shoemake1985animating,
  title     = {Animating rotation with quaternion curves},
  author    = {Shoemake, Ken},
  booktitle = {Proceedings of the 12th annual conference on Computer graphics and interactive techniques},
  pages     = {245--254},
  year      = {1985}
}

@article{skorokhodov2025improving,
  title   = {Improving the diffusability of autoencoders},
  author  = {Skorokhodov, Ivan and Girish, Sharath and Hu, Benran and Menapace, Willi and Li, Yanyu and Abdal, Rameen and Tulyakov, Sergey and Siarohin, Aliaksandr},
  journal = {arXiv preprint arXiv:2502.14831},
  year    = {2025}
}

@inproceedings{song2020score,
  author    = {Song, Yang and Sohl-Dickstein, Jascha and Kingma, Diederik P and Kumar, Abhishek and Ermon, Stefano and Poole, Ben},
  booktitle = {International Conference on Learning Representations},
  title     = {Score-based generative modeling through stochastic differential equations},
  year      = {2021}
}

@book{vershynin2018high,
  title     = {High-dimensional probability: An introduction with applications in data science},
  author    = {Vershynin, Roman},
  volume    = {47},
  year      = {2018},
  publisher = {Cambridge university press}
}

@inproceedings{wang2017normface,
  title     = {Normface: L2 hypersphere embedding for face verification},
  author    = {Wang, Feng and Xiang, Xiang and Cheng, Jian and Yuille, Alan Loddon},
  booktitle = {Proceedings of the 25th ACM international conference on Multimedia},
  pages     = {1041--1049},
  year      = {2017}
}

@inproceedings{xiong2025gigatok,
  title     = {Gigatok: Scaling visual tokenizers to 3 billion parameters for autoregressive image generation},
  author    = {Xiong, Tianwei and Liew, Jun Hao and Huang, Zilong and Feng, Jiashi and Liu, Xihui},
  booktitle = {Proceedings of the IEEE/CVF International Conference on Computer Vision},
  pages     = {18770--18780},
  year      = {2025}
}

@inproceedings{xu2018spherical,
  title     = {Spherical latent spaces for stable variational autoencoders},
  author    = {Xu, Jiacheng and Durrett, Greg},
  booktitle = {Proceedings of the 2018 conference on empirical methods in natural language processing},
  pages     = {4503--4513},
  year      = {2018}
}

@article{xu2026making,
  title   = {Making Reconstruction FID Predictive of Diffusion Generation FID},
  author  = {Xu, Tongda and He, Mingwei and Abu-Hussein, Shady and Hernandez-Lobato, Jose Miguel and Zhang, Haotian and Zhao, Kai and Zhou, Chao and Zhang, Ya-Qin and Wang, Yan},
  journal = {arXiv preprint arXiv:2603.05630},
  year    = {2026}
}

@inproceedings{yao2025reconstruction,
  title     = {Reconstruction vs. generation: Taming optimization dilemma in latent diffusion models},
  author    = {Yao, Jingfeng and Yang, Bin and Wang, Xinggang},
  booktitle = {Proceedings of the IEEE/CVF Conference on Computer Vision and Pattern Recognition},
  year      = {2025}
}

@article{yu2024representation,
  title   = {Representation alignment for generation: Training diffusion transformers is easier than you think},
  author  = {Yu, Sihyun and Kwak, Sangkyung and Jang, Huiwon and Jeong, Jongheon and Huang, Jonathan and Shin, Jinwoo and Xie, Saining},
  journal = {arXiv preprint arXiv:2410.06940},
  year    = {2024}
}

@article{yue2026image,
  title   = {Image Generation with a Sphere Encoder},
  author  = {Yue, Kaiyu and Jia, Menglin and Hou, Ji and Goldstein, Tom},
  journal = {arXiv preprint arXiv:2602.15030},
  year    = {2026}
}

@article{zaghen2025riemannian,
  title   = {Riemannian Variational Flow Matching for Material and Protein Design},
  author  = {Zaghen, Olga and Eijkelboom, Floor and Pouplin, Alison and Liu, Cong and Welling, Max and van de Meent, Jan-Willem and Bekkers, Erik J},
  journal = {arXiv preprint arXiv:2502.12981},
  year    = {2025}
}

@inproceedings{zhang2018unreasonable,
  title     = {The unreasonable effectiveness of deep features as a perceptual metric},
  author    = {Zhang, Richard and Isola, Phillip and Efros, Alexei A and Shechtman, Eli and Wang, Oliver},
  booktitle = {Proceedings of the IEEE conference on computer vision and pattern recognition},
  pages     = {586--595},
  year      = {2018}
}

@article{zheng2025diffusion,
  title   = {Diffusion transformers with representation autoencoders},
  author  = {Zheng, Boyang and Ma, Nanye and Tong, Shengbang and Xie, Saining},
  journal = {arXiv preprint arXiv:2510.11690},
  year    = {2025}
}

\newpage
\appendix
\begin{center}
	{\Large\bfseries Supplementary Material}
\end{center}

\section{Analytical Derivations}

\subsection{Analytical Gaussian Norm Statistics}
\label{app:gaussian-radius}

In \cref{sec:geometry-problem}, we use the standard fact that high-dimensional Gaussian samples concentrate near a spherical shell. Here, we give the analytical norm statistics used for the Gaussian rows in \cref{tab:shell-stats}.

Let
\[
z \sim \mathcal{N}(0,I_d),
\qquad
R = \|z\|_2 .
\]
Because the entries of \(z\) are independent standard normal variables, we have
\[
R^2 = \sum_{i=1}^d z_i^2 \sim \chi_d^2.
\]
Therefore, the norm \(R=\|z\|_2\) follows a chi distribution with \(d\) degrees of freedom.

The exact mean of a chi random variable with \(d\) degrees of freedom is
\[
\mathbb{E}[R]
=
\sqrt{2}\,
\frac{\Gamma((d+1)/2)}{\Gamma(d/2)}.
\]
This is the value we use as the analytical Gaussian mean radius in \cref{tab:shell-stats}.

The second moment is
\[
\mathbb{E}[R^2]
=
\mathbb{E}\|z\|_2^2
=
\sum_{i=1}^d \mathbb{E}[z_i^2]
=
d.
\]
Thus, the variance of the Gaussian norm is
\[
\operatorname{Var}(R)
=
\mathbb{E}[R^2] - \mathbb{E}[R]^2
=
d
-
\left(
\sqrt{2}\,
\frac{\Gamma((d+1)/2)}{\Gamma(d/2)}
\right)^2 .
\]
The coefficient of variation is therefore
\[
\operatorname{CV}(R)
=
\frac{\sqrt{\operatorname{Var}(R)}}{\mathbb{E}[R]}.
\]
These two quantities, \(\mathbb{E}[R]\) and \(\operatorname{CV}(R)\), give the Gaussian entries reported in \cref{tab:shell-stats}.

\paragraph{Relation to the \(\sqrt d\) Gaussian shell.}

The Gaussian shell is often described as being located near radius \(\sqrt d\). This comes from the root-mean-square norm:
\[
\sqrt{\mathbb{E}\|z\|_2^2}
=
\sqrt d.
\]
However, the mean norm is slightly smaller:
\[
\mathbb{E}\|z\|_2 < \sqrt{\mathbb{E}\|z\|_2^2}
=
\sqrt d,
\]
where the inequality follows from Jensen's inequality.

For large \(d\), the exact mean has the expansion
\[
\mathbb{E}\|z\|_2
=
\sqrt d
\left(
1-\frac{1}{4d}
+
O(d^{-2})
\right).
\]
Equivalently,
\[
\mathbb{E}\|z\|_2
=
\sqrt d
-
\frac{1}{4\sqrt d}
+
O(d^{-3/2}).
\]
Thus, the difference between \(\mathbb{E}\|z\|_2\) and \(\sqrt d\) becomes negligible in high dimensions.

Squaring this expansion gives
\[
\left(\mathbb{E}\|z\|_2\right)^2
=
d - \tfrac12 + O(d^{-1}),
\]
which motivates the closed-form approximation
\[
\mathbb{E}\|z\|_2
\approx
\sqrt{d-\tfrac12},
\]
accurate to \(O(d^{-3/2})\). This approximation is accurate at the dimensions used in our experiments. For example,
\[
d=16:
\qquad
\sqrt{d-\frac12}=3.937,
\qquad
\mathbb{E}\|z\|_2=3.938,
\]
and
\[
d=32:
\qquad
\sqrt{d-\frac12}=5.612,
\qquad
\mathbb{E}\|z\|_2=5.613.
\]

Therefore, \(\sqrt d\) should be understood as the conventional approximate shell radius used in concentration results, while the Gaussian rows in \cref{tab:shell-stats} report the exact mean norm \(\mathbb{E}\|z\|_2\).

Finally, the concentration statement in the main paper remains consistent with this calculation. For \(z\sim\mathcal{N}(0,I_d)\),
\[
\mathbb{P}
\left(
\left|
\|z\|_2-\sqrt d
\right|
> t
\right)
\le
2\exp(-c t^2),
\]
for an absolute constant \(c>0\) \citep[Theorem~3.1.1]{vershynin2018high}. This means that Gaussian samples lie in an \(\mathcal{O}(1)\)-width shell around radius \(\sqrt d\). Since the radius itself grows as \(\sqrt d\), the relative shell thickness decreases as \(\mathcal{O}(1/\sqrt d)\).

\subsection{Projected Gaussian Noise is Uniform on the Sphere}
\label{app:projected-gaussian-uniform}

The spherical flow uses a noise endpoint on the fixed-radius sphere \(\mathcal{S}^{d-1}(\sqrt d)\). This prior is the radial projection of the same isotropic Gaussian noise used by the Euclidean baseline.

Let \(\epsilon \sim \mathcal{N}(0, I_d)\) and define
\[
\pi_R(\epsilon) = R\,\frac{\epsilon}{\|\epsilon\|_2}, \qquad R > 0.
\]
Since \(\mathbb{P}(\epsilon = 0) = 0\), this is well-defined almost surely. We show that
\[
\pi_R(\epsilon) \sim \mathrm{Uniform}(\mathcal{S}^{d-1}(R)).
\]

In polar coordinates, write \(\epsilon = r u\) with \(r = \|\epsilon\|_2 \in [0, \infty)\) and \(u \in \mathcal{S}^{d-1}\). The Lebesgue measure factorizes as \(d\epsilon = r^{d-1}\,dr\,d\sigma(u)\), where \(d\sigma\) is the surface measure on \(\mathcal{S}^{d-1}\); after normalization, \(d\sigma/\sigma(\mathcal{S}^{d-1})\) is the uniform probability measure. The Gaussian density therefore satisfies
\[
p(\epsilon)\,d\epsilon = (2\pi)^{-d/2} \exp\!\left(-\frac{r^2}{2}\right) r^{d-1}\,dr \cdot d\sigma(u),
\]
which factorizes into a radial term and a part constant in \(u\). Hence \(u = \epsilon/\|\epsilon\|_2\) is uniform on \(\mathcal{S}^{d-1}\) and independent of \(r\), and
\[
\pi_R(\epsilon) = R u \sim \mathrm{Uniform}(\mathcal{S}^{d-1}(R)) \qquad \text{for every } R > 0.
\]
Setting \(R = \sqrt d\) gives the spherical noise endpoint used in the main paper.

For a latent tensor with \(N = h\cdot w\) spatial positions, let \(\{\epsilon_{i,j}\}\) be \emph{independent} with \(\epsilon_{i,j} \sim \mathcal{N}(0, I_d)\). The token-wise projection is a measurable coordinatewise map, so independence is preserved and the full noise tensor
\[
z_0 = \left\{ \sqrt d\,\frac{\epsilon_{i,j}}{\|\epsilon_{i,j}\|_2} \right\}_{i,j}
\]
is sampled from the product measure \(\bigotimes_{i,j} \mathrm{Uniform}(\mathcal{S}^{d-1}(\sqrt d))\). The spherical prior keeps the Gaussian direction and discards only its radius.

\subsection{Exponential-Map Integration for Slerp Targets}
\label{app:expmap}

This section derives the one-step identity stated in \cref{sec:transport}: along a slerp path, an exponential-map step with the true velocity reaches the exact slerp endpoint, while an Euler step followed by radial projection stays on the same great circle but undershoots by $\sqrt{d}\,\bigl(\Delta t\,\omega - \arctan(\Delta t\,\omega)\bigr)$ in arc length.

We write $R$ for the sphere radius, $\theta_0$ for the endpoint angle, and $h \in [0,\, 1-t]$ for the step size. These correspond to $\sqrt{d}$, $\omega$, and $\Delta t$ in \cref{sec:transport}.

\paragraph{Setup.}
Let $z_0, z_1 \in \mathcal{S}^{d-1}(R)$ with $\langle z_0, z_1\rangle = R^2 \cos\theta_0$ and $\theta_0 \in (0, \pi)$. The slerp path is
\[
z_t \;=\; \frac{\sin((1-t)\theta_0)}{\sin\theta_0}\, z_0 \;+\; \frac{\sin(t\theta_0)}{\sin\theta_0}\, z_1, \qquad t \in [0,1],
\]
which traces the great circle from $z_0$ to $z_1$ at constant angular speed $\theta_0$. Differentiating in $t$ gives the velocity
\[
\dot z_t \;=\; \frac{\theta_0}{\sin\theta_0}\Bigl(-\cos((1-t)\theta_0)\, z_0 \;+\; \cos(t\theta_0)\, z_1\Bigr),
\]
which is tangent to the sphere at $z_t$ and has constant norm $\|\dot z_t\| = R\,\theta_0$.

\paragraph{Exponential-map step.}
At a point $p \in \mathcal{S}^{d-1}(R)$, the exponential map in the direction of a tangent vector $v \in T_p\mathcal{S}^{d-1}(R)$ is
\[
\exp_p(v) \;=\; \cos\!\Bigl(\tfrac{\|v\|}{R}\Bigr)\, p \;+\; R \sin\!\Bigl(\tfrac{\|v\|}{R}\Bigr)\,\frac{v}{\|v\|}.
\]
Stepping from $z_t$ along the true velocity with step size $h$ uses $\|h\dot z_t\| = h R \theta_0$, so
\[
\exp_{z_t}(h\dot z_t) \;=\; \cos(h\theta_0)\, z_t \;+\; \sin(h\theta_0)\,\frac{\dot z_t}{\theta_0}.
\]
Substituting the slerp expressions for $z_t$ and $\dot z_t$ and applying the angle-addition identities for $\sin$ collapses the right-hand side to
\[
\exp_{z_t}(h\dot z_t) \;=\; \frac{\sin((1-t-h)\theta_0)}{\sin\theta_0}\, z_0 \;+\; \frac{\sin((t+h)\theta_0)}{\sin\theta_0}\, z_1 \;=\; z_{t+h},
\]
the slerp value at time $t+h$. With a perfect velocity predictor, the exponential-map sampler is therefore exact along the slerp curve.

\paragraph{Euler step followed by radial projection.}
The Euler step produces $\tilde z = z_t + h\, \dot z_t$, which lies in the plane spanned by $\{z_t,\, \dot z_t/\theta_0\}$, the same plane that contains the slerp great circle. Since $z_t \perp \dot z_t$ and $\|\dot z_t\| = R\theta_0$, its squared norm is
\[
\|\tilde z\|^2 \;=\; R^2 + h^2 R^2 \theta_0^2 \;=\; R^2\bigl(1 + h^2\theta_0^2\bigr).
\]
Radial projection $\Pi(\tilde z) = R\,\tilde z/\|\tilde z\|$ then gives
\[
\Pi(\tilde z) \;=\; \frac{1}{\sqrt{1+h^2\theta_0^2}}\, z_t \;+\; \frac{h\theta_0}{\sqrt{1+h^2\theta_0^2}}\,\frac{\dot z_t}{\theta_0}.
\]
Matching this against the exponential-map form $\cos\phi\, z_t + \sin\phi\, \dot z_t/\theta_0$ identifies the projected Euler step as a rotation of $z_t$ on the same great circle, but at the smaller angular displacement $\arctan(h\theta_0)$ rather than $h\theta_0$.

\paragraph{One-step deficit.}
Both samplers stay on the great circle through $z_t$ and $z_{t+h}$. The exponential-map step advances by arc length $R\, h\theta_0$, while the projected Euler step advances by $R\,\arctan(h\theta_0)$. The one-step arc-length deficit is therefore
\[
R\bigl(h\theta_0 - \arctan(h\theta_0)\bigr) \;=\; R\,\frac{(h\theta_0)^3}{3} \;+\; O\bigl((h\theta_0)^5\bigr),
\]
which reduces to the expression quoted in \cref{sec:transport} after substituting $R = \sqrt{d}$, $\theta_0 = \omega$, $h = \Delta t$.

\section{Implementation Details}

\subsection{Slerp Numerical Handling}
\label{sec:slerp-numerical}

Per-token slerp treats each spatial position's $d$-dimensional vector as a point on $\mathcal{S}^{d-1}(\sqrt{d})$. The implementation handles three regimes by the angle $\omega$ between the unit-normalized endpoints. For $\omega \in [10^{-4},\, \pi - 0.1]$ the standard slerp formula applies. For $\omega < 10^{-4}$ the path falls back to linear interpolation followed by renormalization to avoid $0/0$ in $\sin\omega$. For $\omega > \pi - 0.1$ the path traces $\cos(\pi t)\,\hat{x}_0 + \sin(\pi t)\,\hat{n}$, an arbitrary great circle from $\hat{x}_0$ through an orthogonal unit $\hat{n}$ to $-\hat{x}_0$. Cosines are clamped to $[-1+10^{-6},\,1-10^{-6}]$ before $\arccos$ and norms are floored at $10^{-8}$. For random unit vectors with $d \in \{16, 32\}$, the per-token cosine concentrates near zero with scale $1/\sqrt{d}$, so the small-angle and antipodal branches are correctness insurance rather than hot paths during training.

\subsection{Training Hyperparameters}
\label{sec:hyperparameters}

Hyperparameters are grid-searched at SiT-B with the FLUX.2 tokenizer and applied unchanged at SiT-XL and to the VA-VAE and REPA-E FLUX.1 tokenizers, with only the timestep shift adapted per tokenizer. The linear and spherical recipes share these hyperparameters and differ only in the path (Euclidean linear vs.\ slerp), the latent (raw VAE vs.\ projected to $\mathcal{S}^{d-1}(\sqrt{d})$), and the slerp-specific design choices already documented in \cref{sec:transport} (tangent projection on the model output, exponential-map sampler). \Cref{tab:hp} lists the resulting configuration.

\begin{table}[h]
	\caption{Training hyperparameters used for all linear-baseline and spherical-slerp runs in the main paper. Only the timestep shift varies across tokenizers; all other settings are shared.}
	\label{tab:hp}
	\centering
	\small
	\begin{tabular}{@{}ll@{}}
		\toprule
		Setting              & Value                                                                       \\
		\midrule
		Optimizer            & AdamW \citep{loshchilov2017decoupled}, $\beta_1{=}0.9$, $\beta_2{=}0.999$    \\
		Weight decay         & $0$                                                                         \\
		EMA decay            & $0.9999$, averaged from step $0$, no warmup                                 \\
		Mixed precision      & bf16 (parameters and EMA in fp32; forward/backward autocast)                \\
		Global batch size    & $256$                                                                       \\
		Training duration    & $80$ epochs                                                                  \\
		Learning rate        & $4{\times}10^{-4}$                                                          \\
		LR schedule          & constant, no warmup                                                         \\
		Gradient clip        & $1.0$                                                                       \\
		Time sampling        & logit-normal                                                                \\
		Timestep shift       & $4.63$ (FLUX.2, REPA-E FLUX.1), $1.0$ (VA-VAE)                              \\
		Training-curve FID-50K  & $50$K samples, $50$ sampling steps, CFG$=1.0$, every $50$K steps            \\
		Final FID-50K           & $50$K samples, $250$ sampling steps, CFG values per \cref{tab:cross-vae}    \\
		\bottomrule
	\end{tabular}
\end{table}

\subsection{Compute}
\label{sec:compute}

All experiments run on NVIDIA H200 GPUs. SiT-B runs use 2 GPUs each; SiT-XL runs use 4 GPUs each. Token counts are equalized across tokenizers by pairing the patch size with the VAE downsample factor: SiT-B/1 and XL/1 for VA-VAE (downsample factor $16$), and SiT-B/2 and XL/2 for FLUX.2 and REPA-E FLUX.1 (downsample factor $8$). All settings yield $256$ tokens per $256{\times}256$ image.

A B-scale run completes 80 epochs in approximately $6.5$ wall-clock hours on $2{\times}$H200 ($\approx 13$ H200-hours); an XL-scale run completes in $\approx 16.8$ hours on $4{\times}$H200 ($\approx 67$ H200-hours).

\subsection{Code and Checkpoint Release}
\label{sec:release}

We will release training and evaluation code, finetuning configurations for the FLUX.2, VA-VAE, and REPA-E FLUX.1 spherical decoders, and the trained SiT-B and SiT-XL flow-matching checkpoints reported in \cref{tab:cross-vae}.

\section{Additional Experiments and Ablations}

\subsection{Core Recipe Checks}
\label{sec:recipe}

Recipe ablations are run on FLUX.2 as the canonical tokenizer; cross-tokenizer transfer of the resulting recipe at B-scale across FLUX.2, VA-VAE, and REPA-E FLUX.1 is reported in the main paper (\cref{tab:cross-vae}). The matched-recipe decoder finetune does not by itself explain the gain in \cref{tab:transport}: applying it to the vanilla decoder under a vanilla-linear flow moves FID only $26.35 \!\to\! 26.95$ (\cref{tab:control}), so the remaining improvement comes from the latent geometry rather than from the finetune.

\begin{table}[h]
	\caption{Selected recipe checks on SiT-B/2 with the FLUX.2 spherical VAE and slerp path (80 epochs, FID-50K (50 steps) at CFG$=1.0$). Each row changes one method-relevant factor from the default.}
	\label{tab:recipe}
	\centering
	\begin{tabular}{lc}
		\toprule
		Configuration                              & FID$\downarrow$ \\
		\midrule
		\rowcolor{gray!20} Spherical-slerp default & 20.55           \\
		Uniform time sampling                      & 23.86           \\
		Unit-radius sphere                         & 22.08           \\
		No tangent projection                      & 25.79           \\
		\bottomrule
	\end{tabular}
\end{table}

\subsection{Recipe Parity}
\label{sec:recipe-parity}

Matching the training recipe improves the vanilla baseline, but the spherical-slerp geometry remains better (\cref{tab:recipe-parity}).

\begin{table}[h]
	\caption{Recipe parity on SiT-B/2 with the FLUX.2 VAE (80 epochs, FID-50K (50 steps) at CFG$=1.0$).}
	\label{tab:recipe-parity}
	\centering
	\begin{tabular}{llc}
		\toprule
		Configuration   & Training recipe & FID$\downarrow$ \\
		\midrule
		Vanilla-linear  & Baseline        & 34.30           \\
		Vanilla-linear  & Matched         & 26.35           \\
		Spherical-slerp & Matched         & 20.55           \\
		\bottomrule
	\end{tabular}
\end{table}

\subsection{Decoder-Flow Coupling}
\label{sec:control}

The gain comes from the spherical projection, not the finetune compute. On FLUX.2 B/2, a matched-recipe decoder finetune \emph{without} the projection remains tied with the vanilla baseline, while swapping decoders between vanilla and spherical flows degrades both pairings (\cref{tab:control}). The flow learns a velocity field calibrated to a specific latent geometry and the decoder learns to invert that geometry; the two co-adapt and cannot be substituted independently.

\begin{table}[h]
	\caption{Decoder-flow coupling on FLUX.2 SiT-B/2 (CFG$=$1.0).}
	\label{tab:control}
	\centering
	\begin{tabular}{@{}lllllr@{}}
		\toprule
		Latent    & Path   & Decoder   & Finetune & Prior  & FID$\downarrow$ \\
		\midrule
		Vanilla     & Linear & Vanilla   & No         & Gaussian & 26.35  \\
		Vanilla     & Linear & Vanilla   & Matched    & Gaussian & 26.95  \\
		Spherical   & Slerp  & Spherical & Matched    & Sph. Uniform  & 20.55  \\
		\midrule
		Vanilla     & Linear & Spherical & Matched    & Gaussian & 112.17 \\
		Spherical   & Slerp  & Vanilla   & No         & Sph. Uniform  & 43.17  \\
		\bottomrule
	\end{tabular}
\end{table}

\subsection{Scaling with Number of Function Evaluations (NFE)}
\label{sec:nfe}

\Cref{tab:nfe} sweeps the number of function evaluations (NFE) on VA-VAE B/1 at CFG $=1.0$. Spherical-slerp wins at every tested NFE, with the gap largest at NFE $=20$ ($-19.84$ FID) and narrowing to $-5.30$ at NFE $=500$; the advantage holds across sampling budgets, not just at low step counts.

\begin{table}[h]
	\caption{Number of function evaluations (NFE) scaling on ImageNet-256 with VA-VAE B/1 at CFG$=1.0$.}
	\label{tab:nfe}
	\centering
	\begin{tabular}{lccc}
		\toprule
		NFE & Vanilla-linear & Spherical-slerp & Gap \\
		\midrule
		20  & 70.44 & 50.60 & $-19.84$ \\
		50  & 31.99 & 23.88 & $-8.11$  \\
		100 & 28.25 & 22.04 & $-6.21$  \\
		500 & 27.26 & 21.96 & $-5.30$  \\
		\bottomrule
	\end{tabular}
\end{table}

\subsection{Direction-vs-Radius Component Ablation}
\label{sec:component-swap-detail}

\begin{figure}[h]
	\centering
	\includegraphics[width=0.85\linewidth]{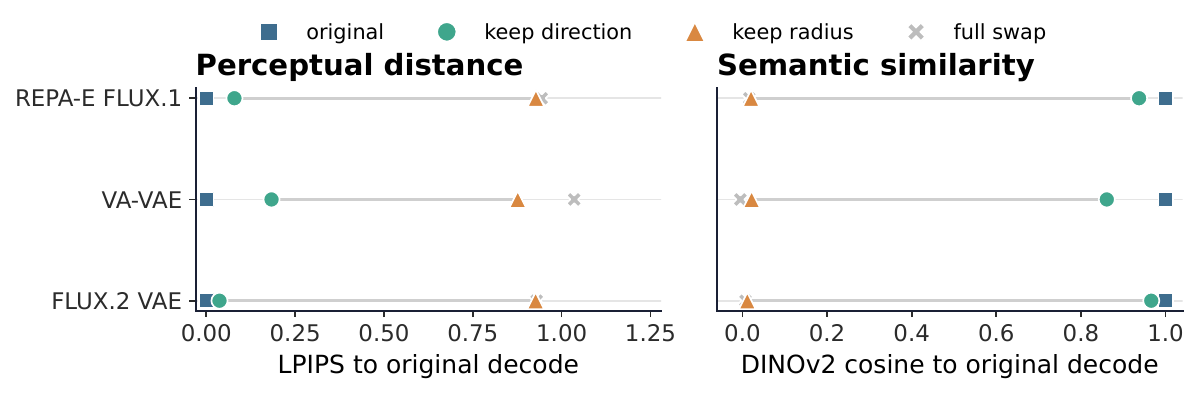}
	\caption{Angular/radial decoder sensitivity, population-mean substitute. For each tokenizer, markers report mean LPIPS (left) and DINOv2 cosine similarity (right) to the original decode under four conditions: original, keep direction, keep radius, full mean substitute; 1024 images per condition.}
	\label{fig:component-ablation-global}
\end{figure}

\Cref{fig:component-ablation} substitutes a same-class partner's direction or radius for the anchor's. As a stronger baseline, \cref{fig:component-ablation-global} substitutes the population-mean direction and radius across the whole ImageNet validation split, isolating whether radius information is carried by a single dataset-wide constant. Each tokenizer reports LPIPS and DINOv2 cosine similarity to the original decode for four conditions: original, keep direction (own direction with substitute radius), keep radius (substitute direction with own radius), and full substitute.

For FLUX.2 and REPA-E FLUX.1 with the population-mean substitute, keep-direction reaches LPIPS $\approx 0.04$--$0.08$ and DINOv2 cosine $\approx 0.94$--$0.97$: replacing the radius with a single global constant leaves the decode visually almost unchanged. VA-VAE is somewhat further (LPIPS $\approx 0.18$, DINOv2 $\approx 0.86$) but still recognizable. Keep-radius lies at LPIPS $\approx 0.88$--$0.93$ and DINOv2 $\approx 0.01$--$0.02$ across all three tokenizers, indistinguishable from the full mean substitute: once the direction is replaced, additionally replacing the radius changes nothing further. The same-class substitute (\cref{fig:component-ablation}) yields the same pattern: keep-direction stays near the original, while keep-radius nearly matches the full-neighbor swap.

\begin{figure}[h]
	\centering
	\includegraphics[width=\textwidth]{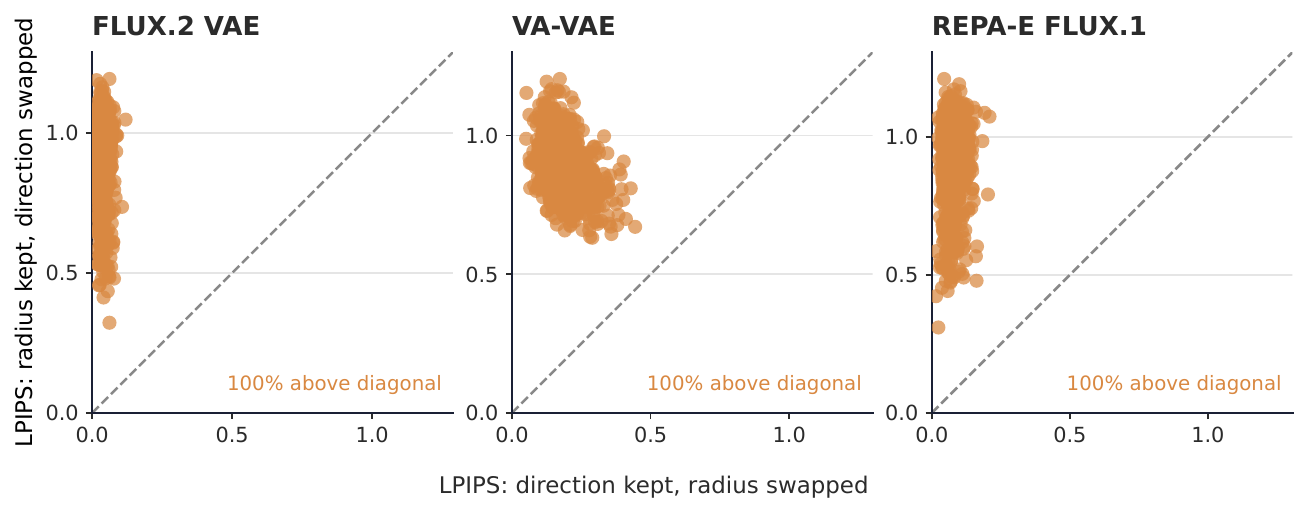}
	\caption{Per-sample direction vs.\ radius sensitivity, population-mean substitute. Each point is one image; x-axis: LPIPS after replacing radius with the population mean (direction kept); y-axis: LPIPS after replacing direction with the population mean (radius kept). Panels: FLUX.2, VA-VAE, REPA-E FLUX.1.}
	\label{fig:perturb}
\end{figure}

\begin{figure}[h]
	\centering
	\includegraphics[width=\textwidth]{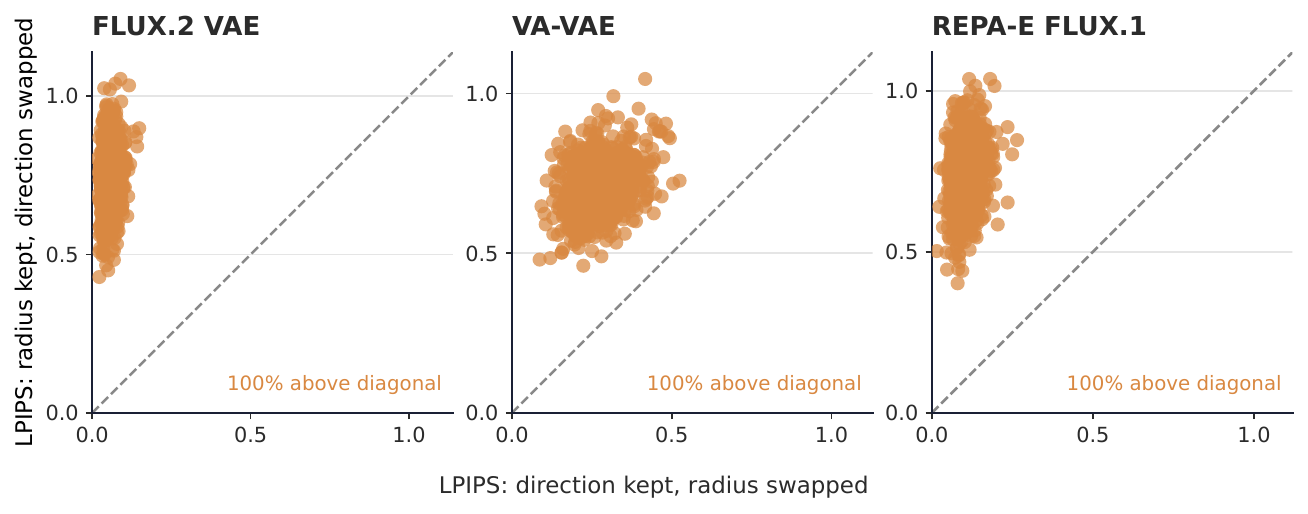}
	\caption{Per-sample direction vs.\ radius sensitivity, same-class partner substitute. Axes as in \cref{fig:perturb}; the substitute is a same-class partner's radius (x-axis) or direction (y-axis).}
	\label{fig:perturb-sameclass}
\end{figure}

\Cref{fig:perturb,fig:perturb-sameclass} plot the per-sample distributions underlying these averages, one point per image, x-axis is LPIPS for keep direction, y-axis is LPIPS for keep radius. Every point lies above $y{=}x$ for all three tokenizers in both substitute regimes; the same-class substitute spreads the radius axis slightly further than the population-mean substitute, most visibly for VA-VAE.

Decoded content is carried almost entirely by the angular direction; the decoder appears largely insensitive to radius. Linear flow paths spend substantial velocity on this radial motion (\cref{fig:flow-radial-energy}); slerp on the sphere has no radial component by construction.

\subsection{Scope of Held-Out Comparisons}
\label{sec:held-out}

REPA-E~\citep{leng2025repa} jointly trains the VAE and the diffusion model under a representation-alignment objective; matching its compute requires retraining both stacks under that objective, which is out of scope for a study that holds the tokenizer fixed and modifies only the latent geometry. \citet{kumar2026learning} train flow matching directly in the frozen DINOv2 feature space, a different tokenizer regime that requires running the encoder at inference. The setting studied here (a geometric constraint on the latent of a standard pretrained VAE, with no auxiliary encoder at training or inference time) uses less compute than these two methods at comparable diffusion-model scale.

\section{Qualitative Results}

\subsection{Class-Conditional Samples}
\label{sec:qualitative-generations}

\begin{figure}[p]
	\centering
	\begin{subfigure}[t]{0.26\textwidth}
		\includegraphics[width=\textwidth]{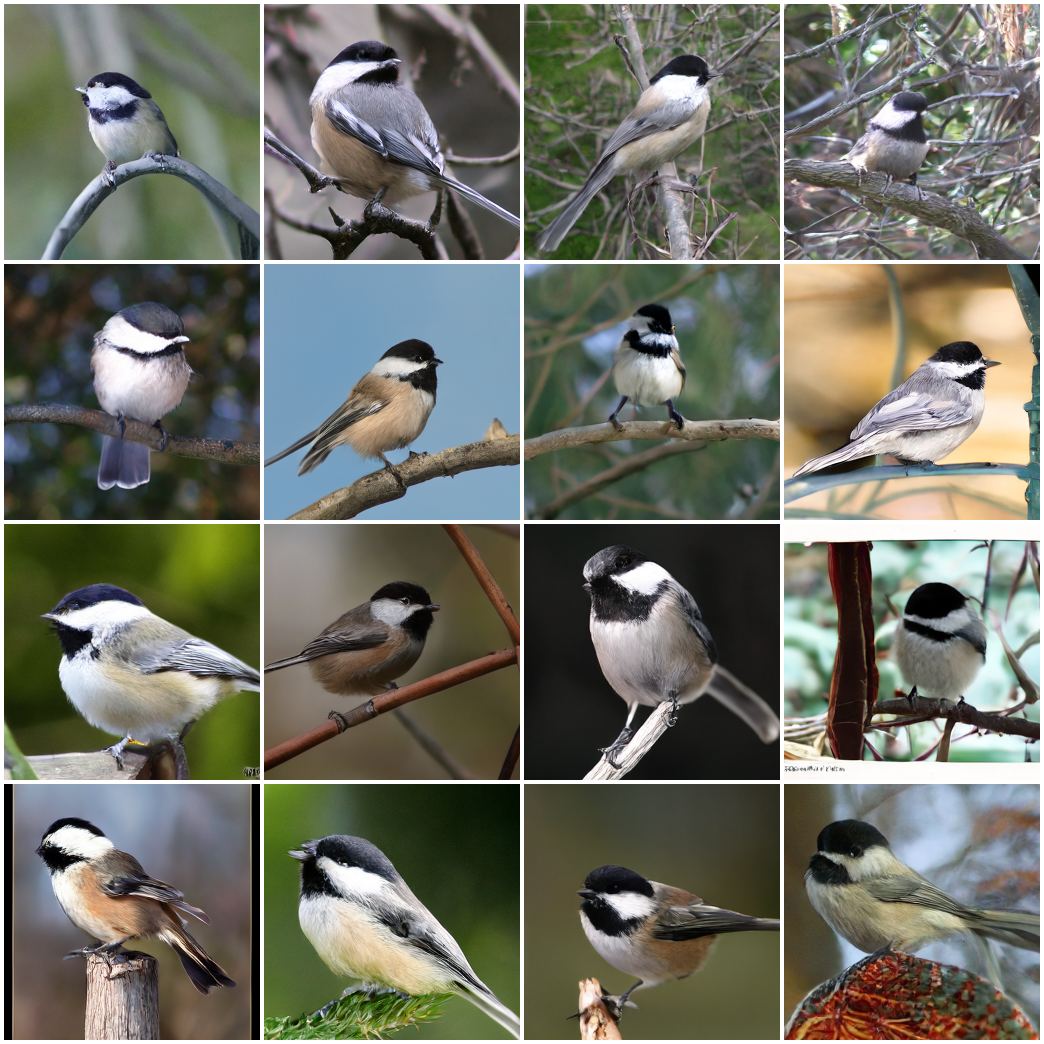}
		\caption{Chickadee (19)}
	\end{subfigure}
	\hfill
	\begin{subfigure}[t]{0.26\textwidth}
		\includegraphics[width=\textwidth]{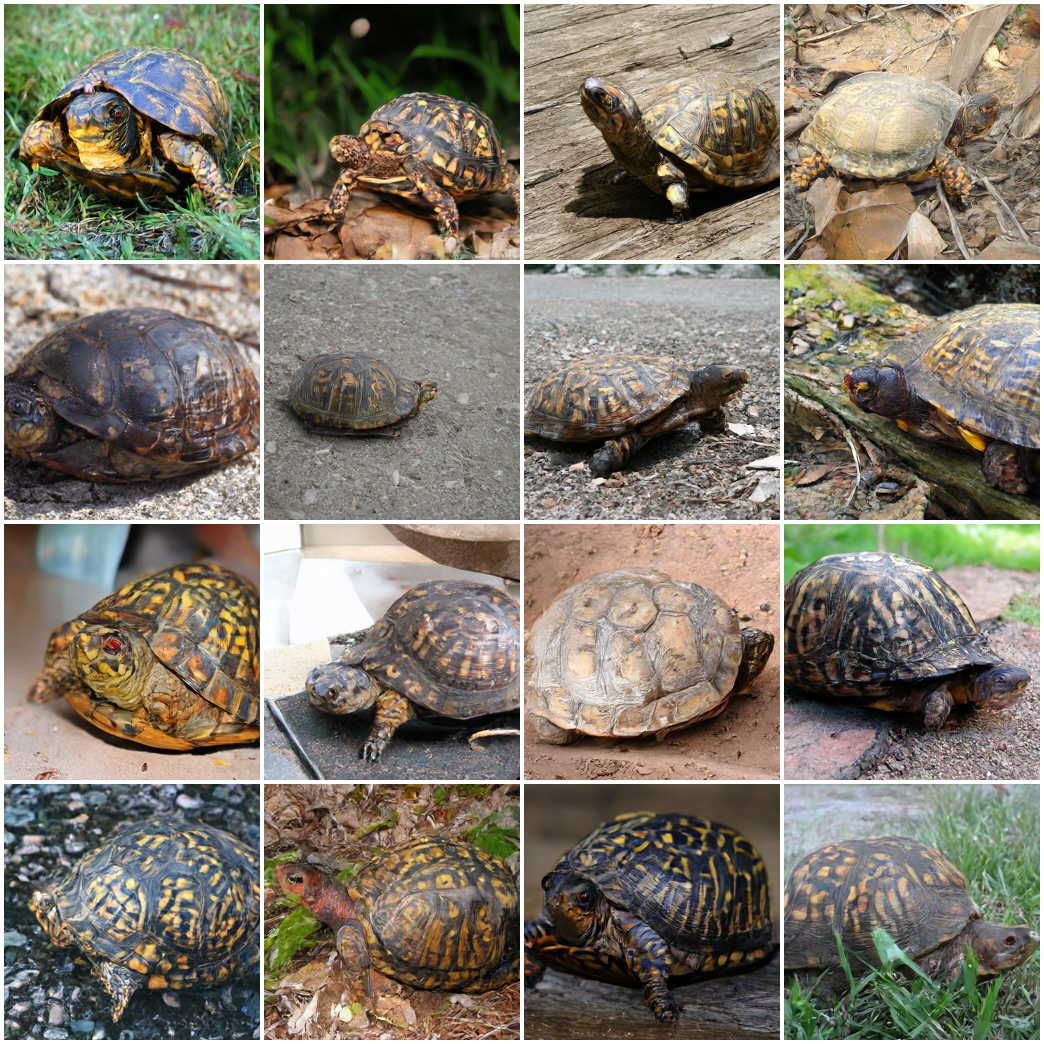}
		\caption{Box turtle (37)}
	\end{subfigure}
	\hfill
	\begin{subfigure}[t]{0.26\textwidth}
		\includegraphics[width=\textwidth]{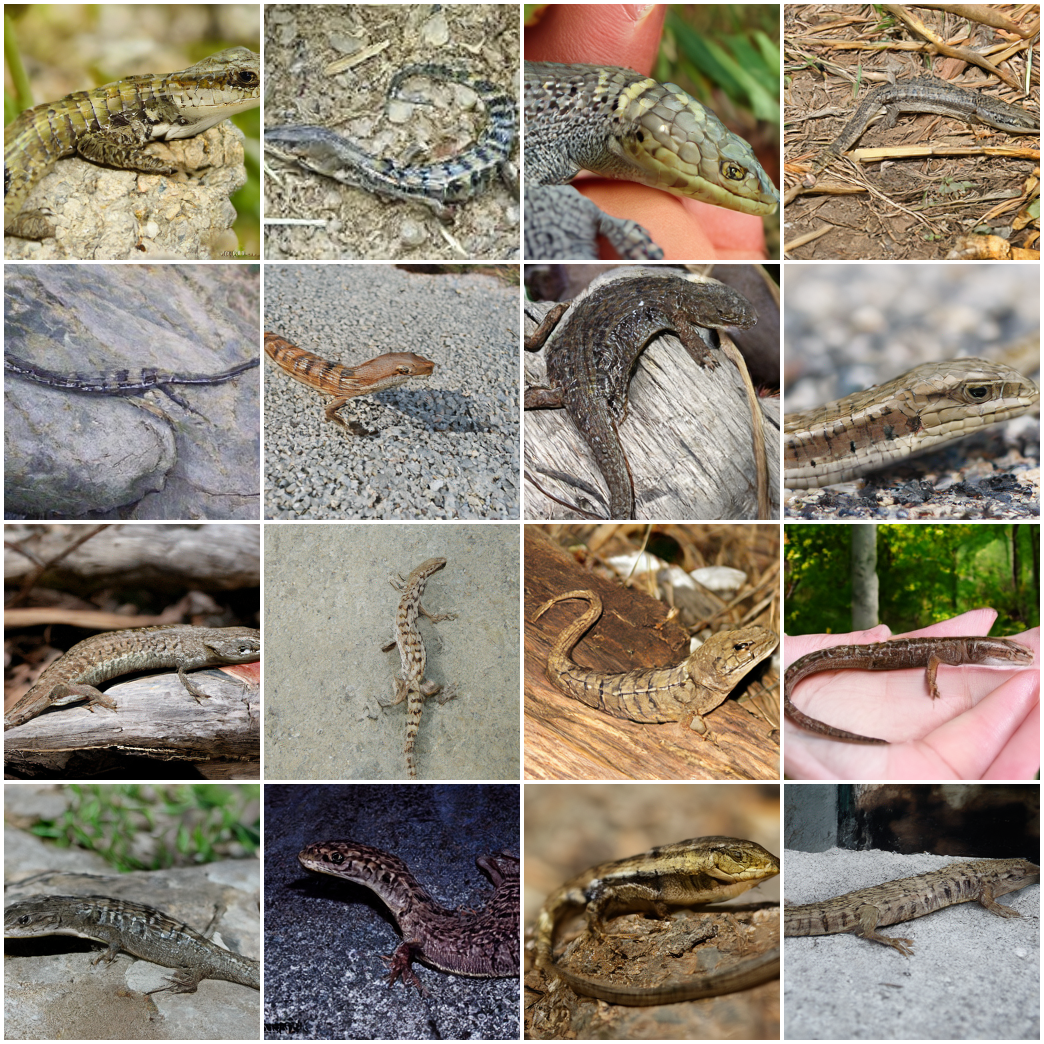}
		\caption{Alligator lizard (44)}
	\end{subfigure}

	\begin{subfigure}[t]{0.26\textwidth}
		\includegraphics[width=\textwidth]{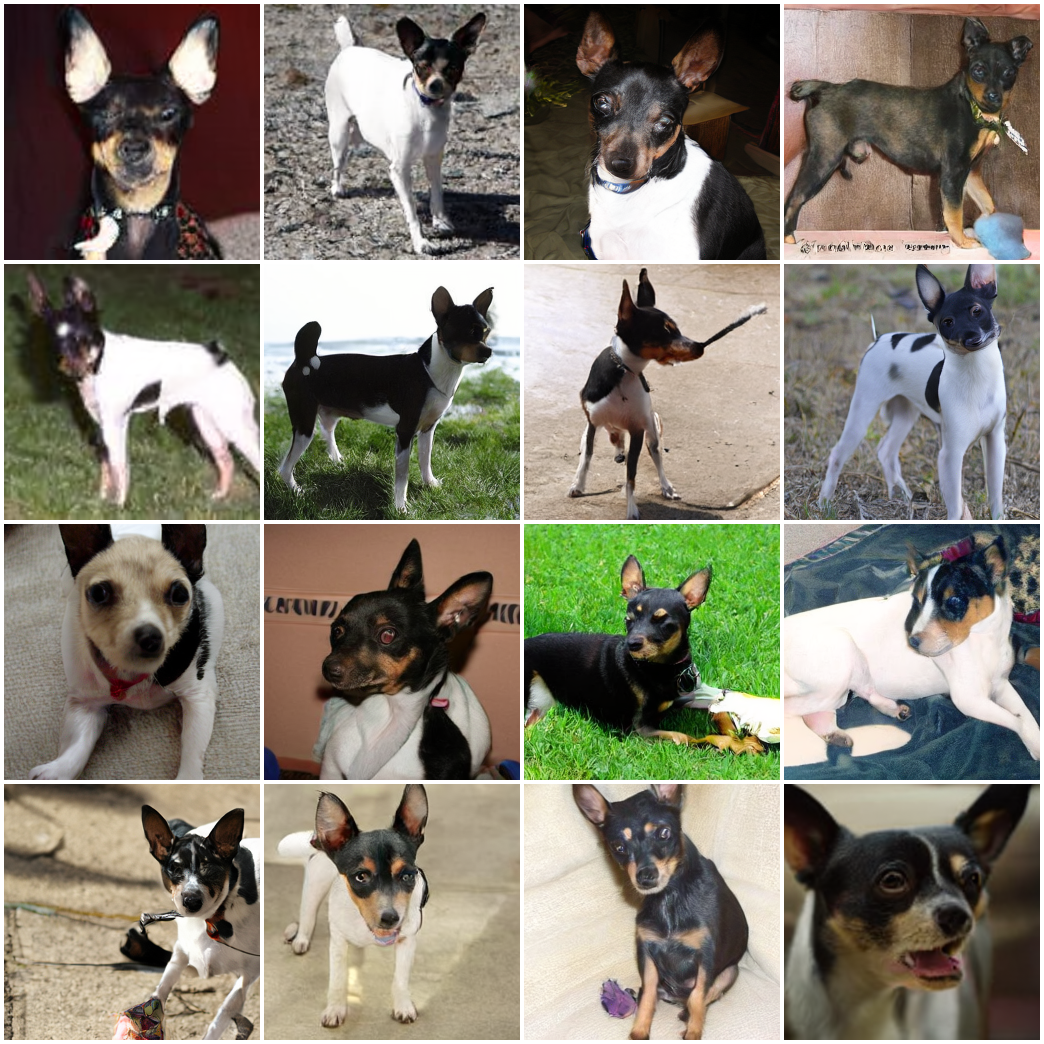}
		\caption{Toy terrier (158)}
	\end{subfigure}
	\hfill
	\begin{subfigure}[t]{0.26\textwidth}
		\includegraphics[width=\textwidth]{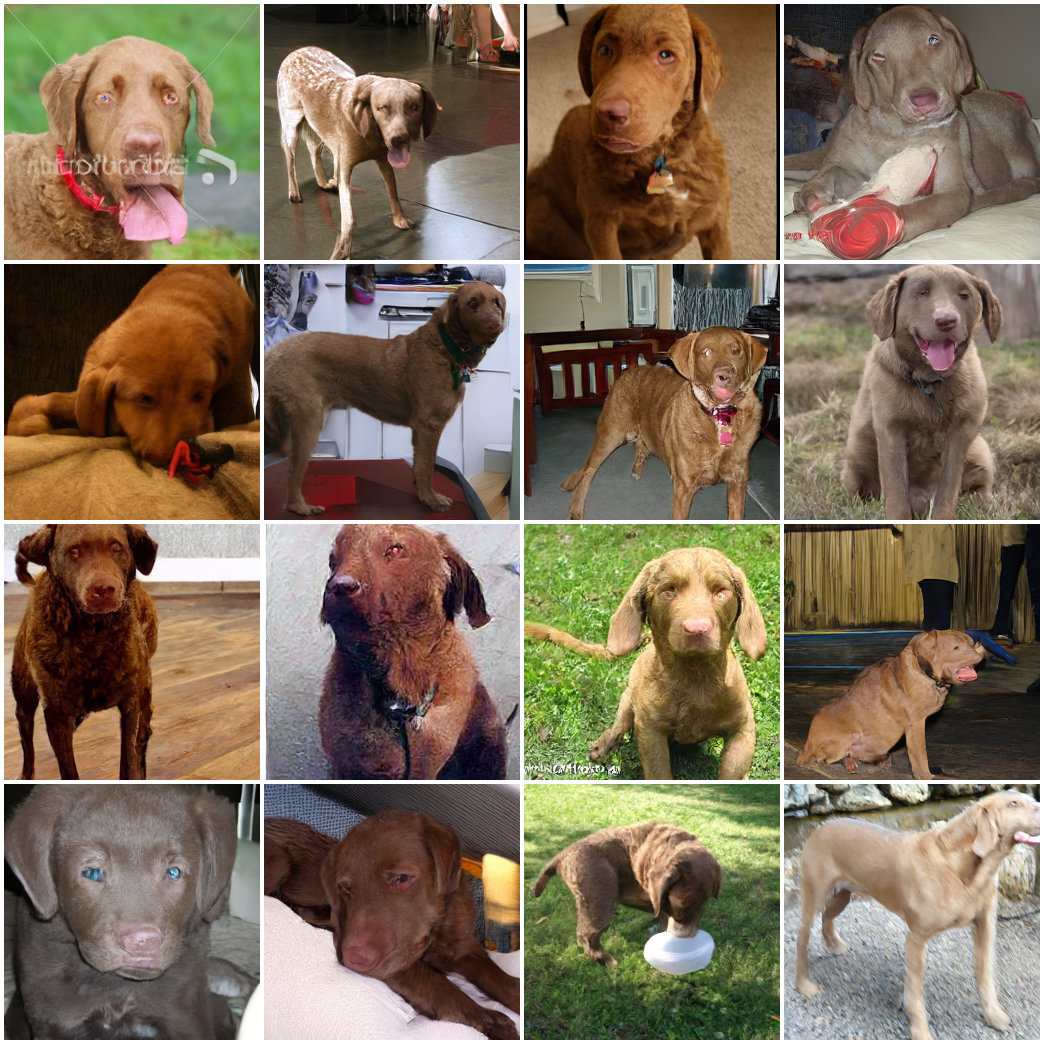}
		\caption{Chesapeake Bay retriever (209)}
	\end{subfigure}
	\hfill
	\begin{subfigure}[t]{0.26\textwidth}
		\includegraphics[width=\textwidth]{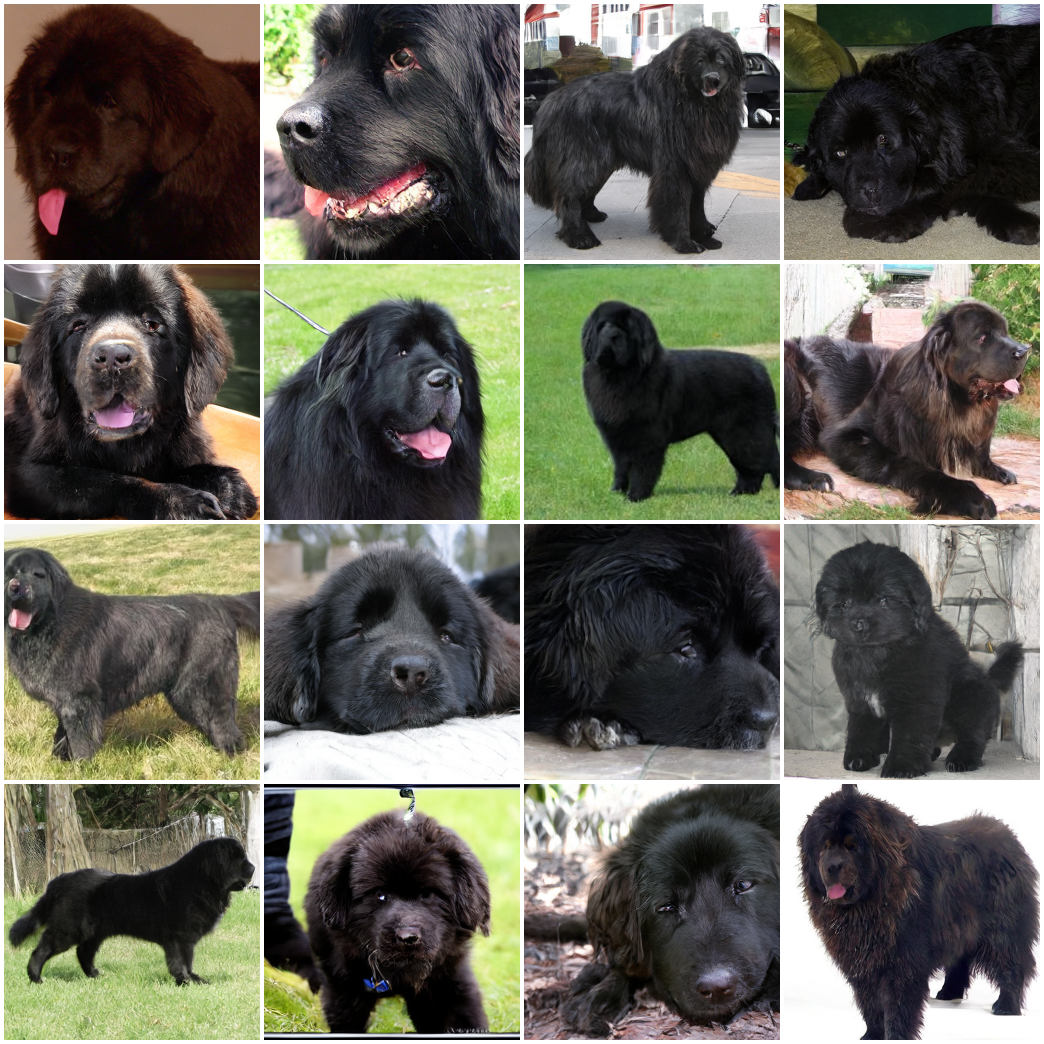}
		\caption{Newfoundland (256)}
	\end{subfigure}

	\begin{subfigure}[t]{0.26\textwidth}
		\includegraphics[width=\textwidth]{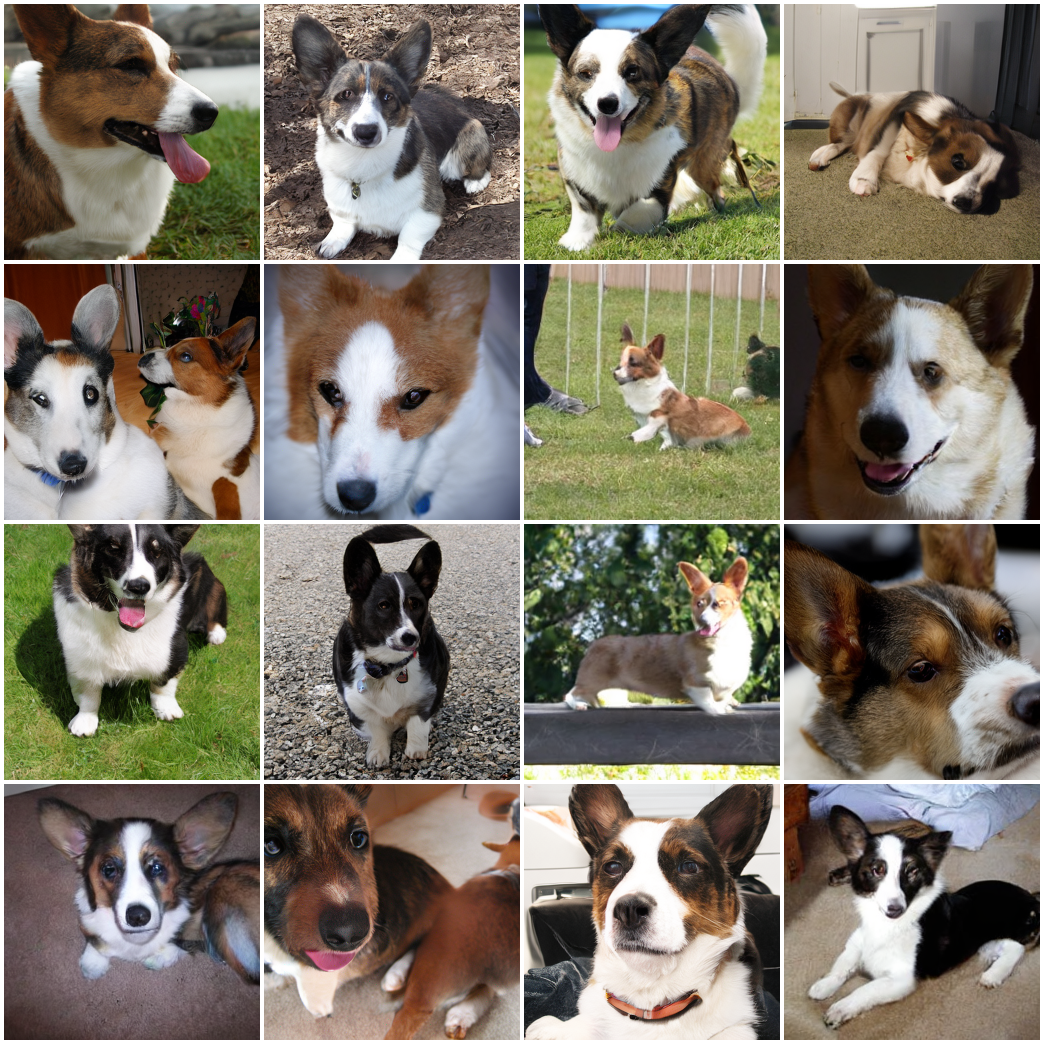}
		\caption{Cardigan (264)}
	\end{subfigure}
	\hfill
	\begin{subfigure}[t]{0.26\textwidth}
		\includegraphics[width=\textwidth]{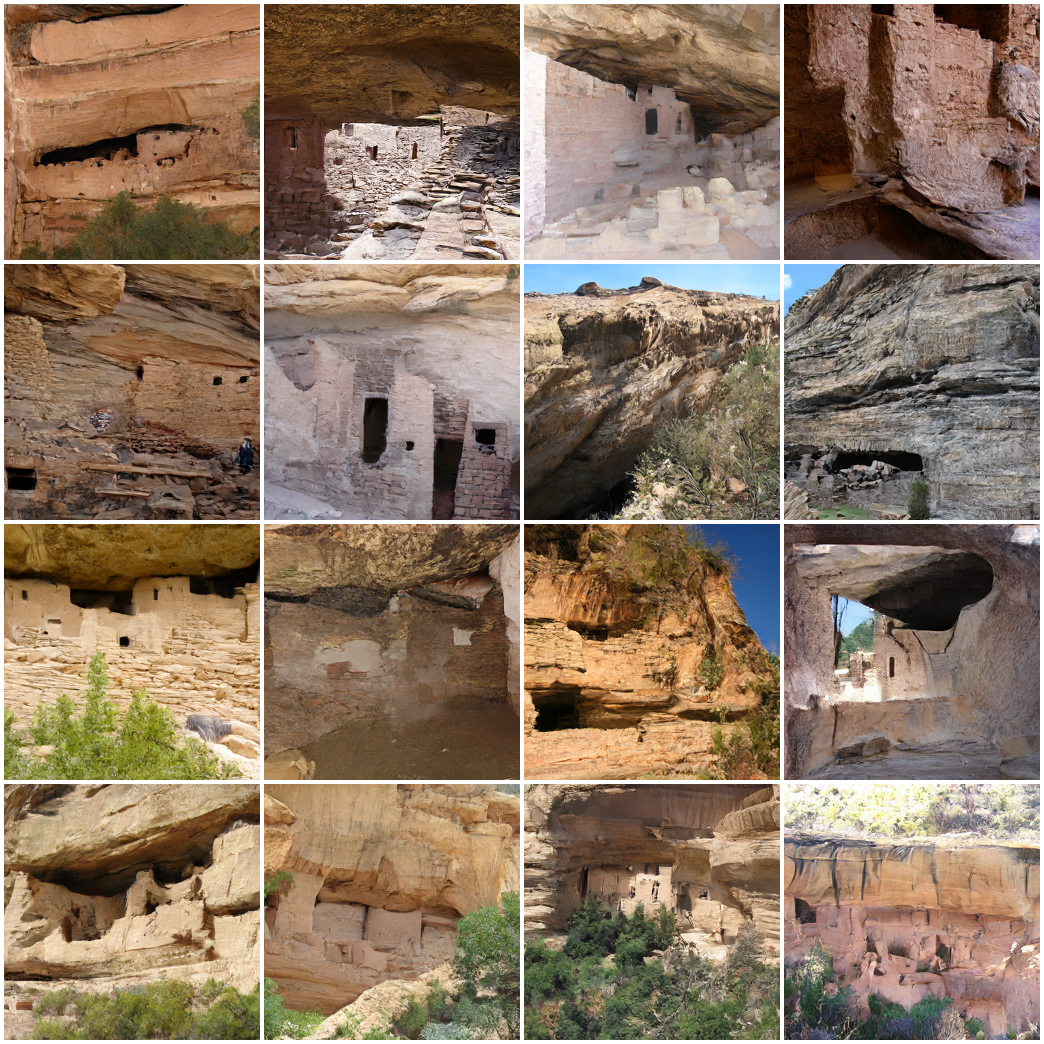}
		\caption{Cliff dwelling (500)}
	\end{subfigure}
	\hfill
	\begin{subfigure}[t]{0.26\textwidth}
		\includegraphics[width=\textwidth]{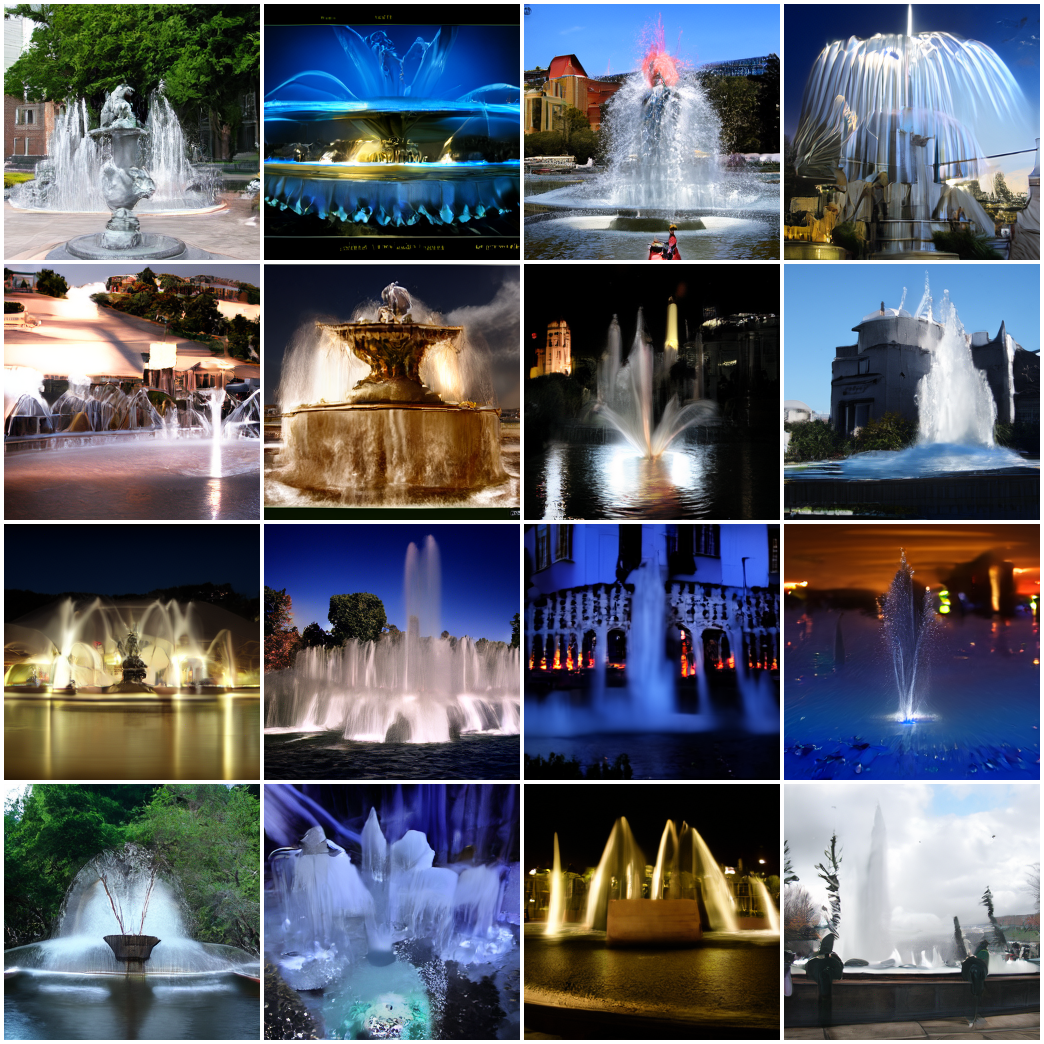}
		\caption{Fountain (562)}
	\end{subfigure}

	\begin{subfigure}[t]{0.26\textwidth}
		\includegraphics[width=\textwidth]{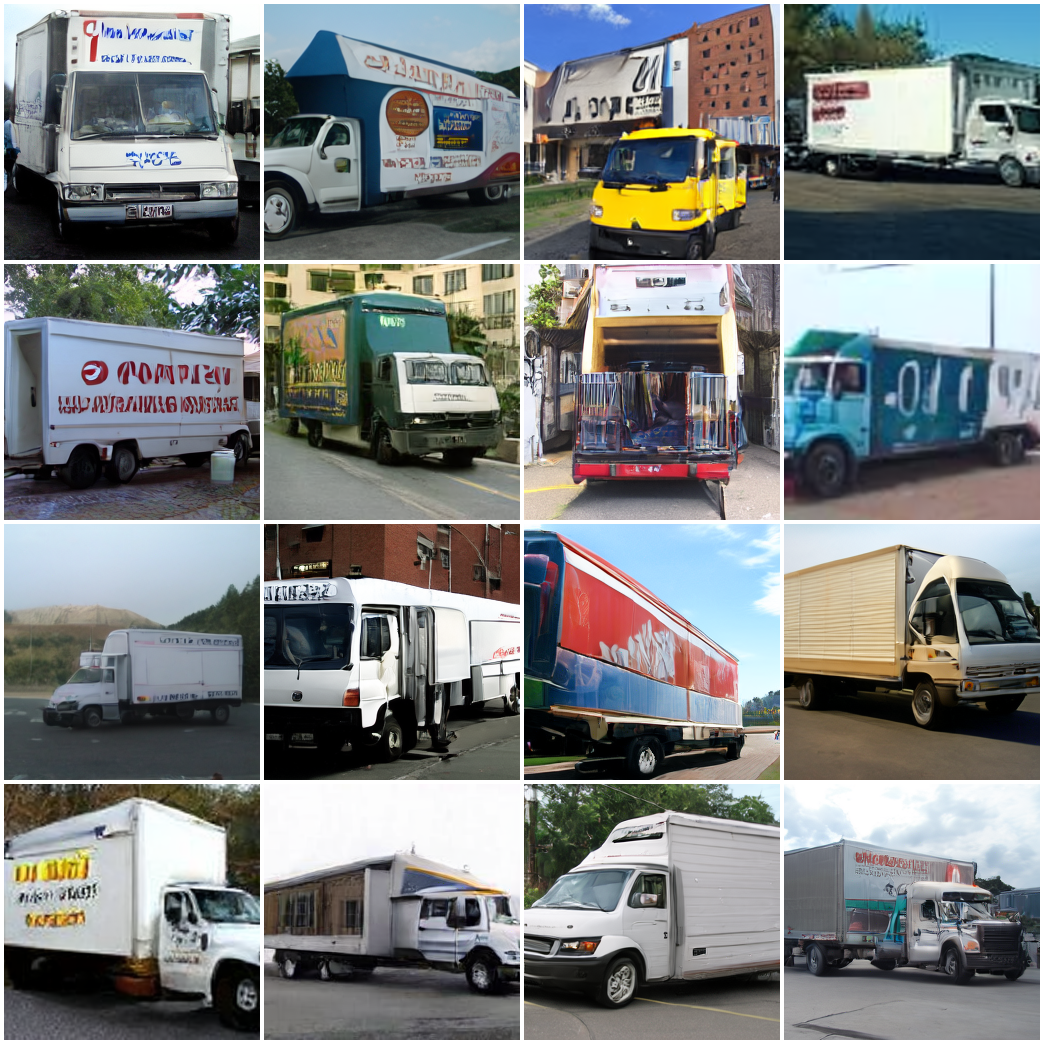}
		\caption{Moving van (675)}
	\end{subfigure}
	\hfill
	\begin{subfigure}[t]{0.26\textwidth}
		\includegraphics[width=\textwidth]{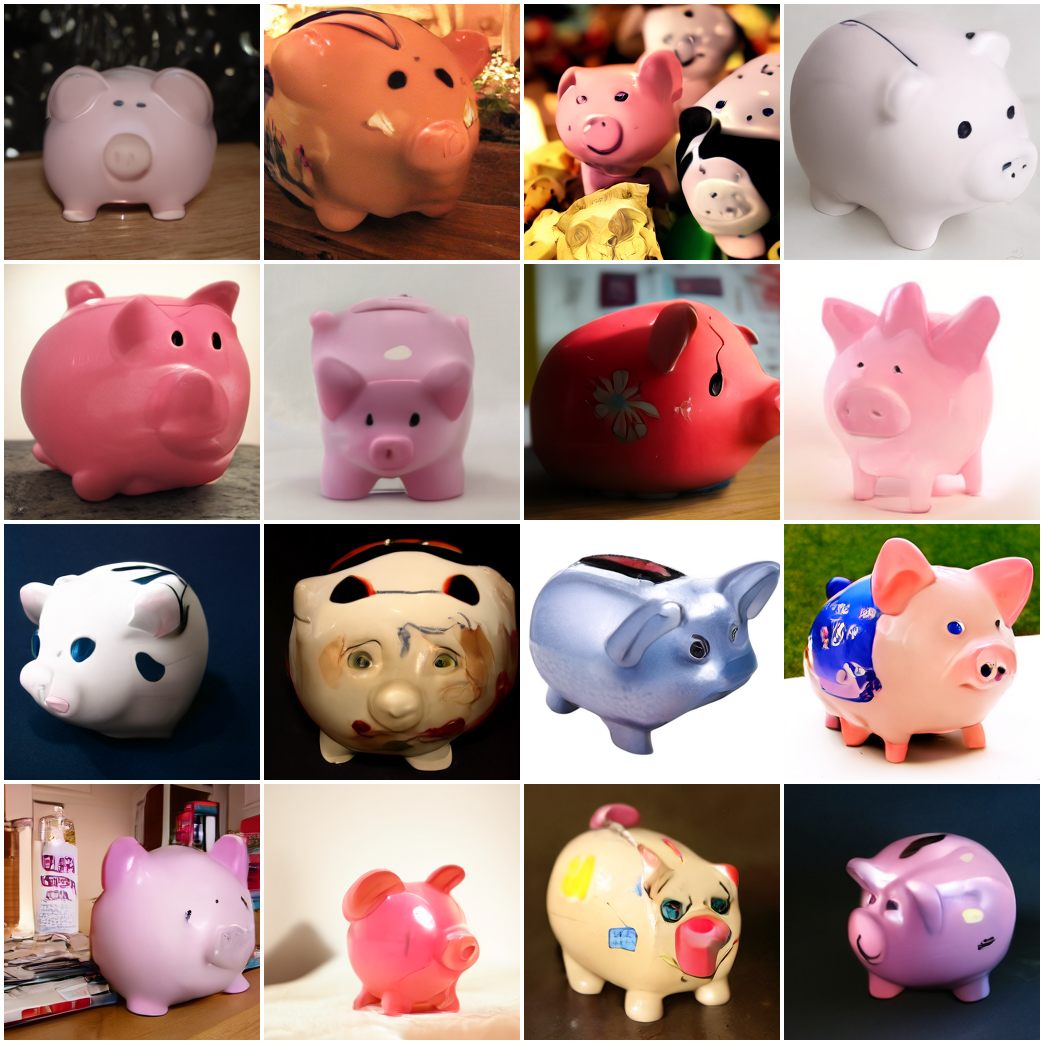}
		\caption{Piggy bank (719)}
	\end{subfigure}
	\hfill
	\begin{subfigure}[t]{0.26\textwidth}
		\includegraphics[width=\textwidth]{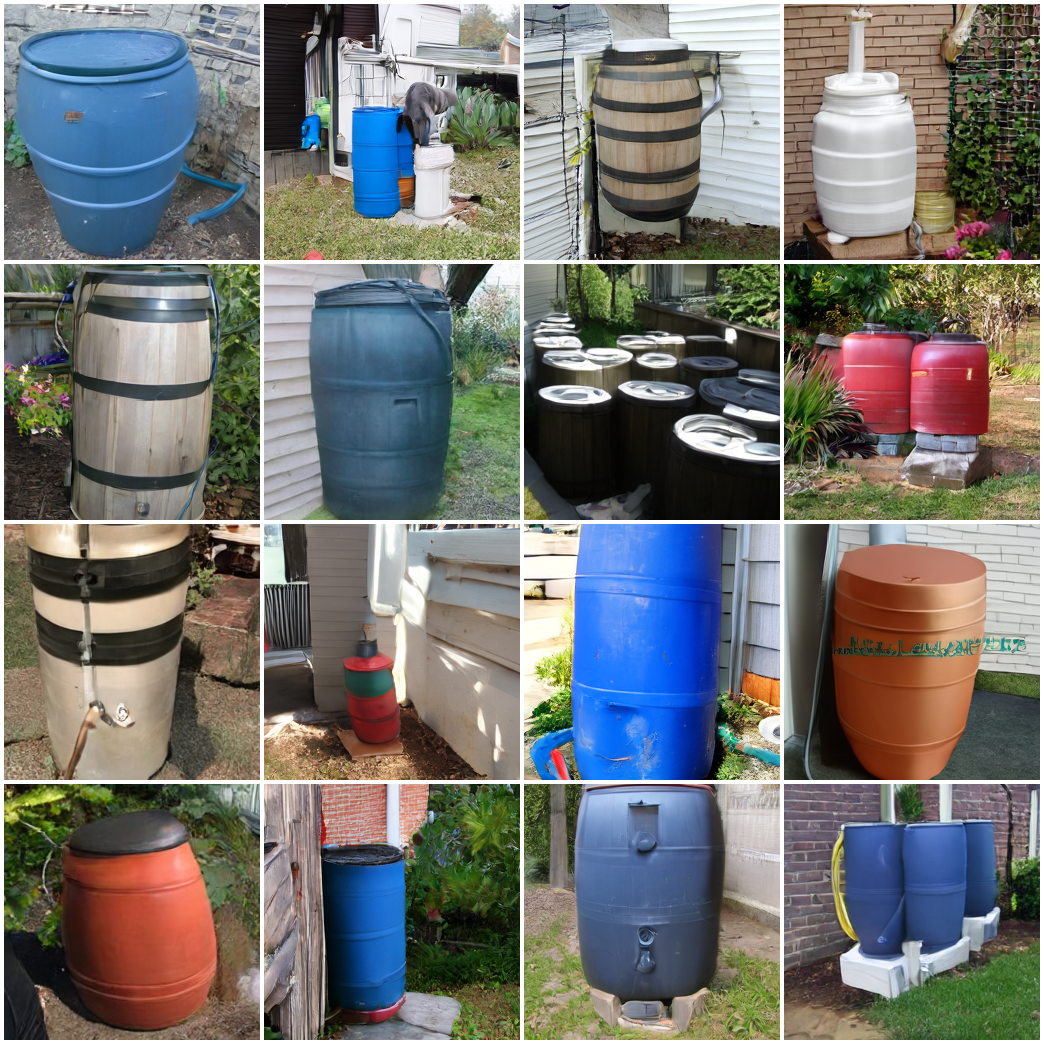}
		\caption{Rain barrel (756)}
	\end{subfigure}

	\begin{subfigure}[t]{0.26\textwidth}
		\includegraphics[width=\textwidth]{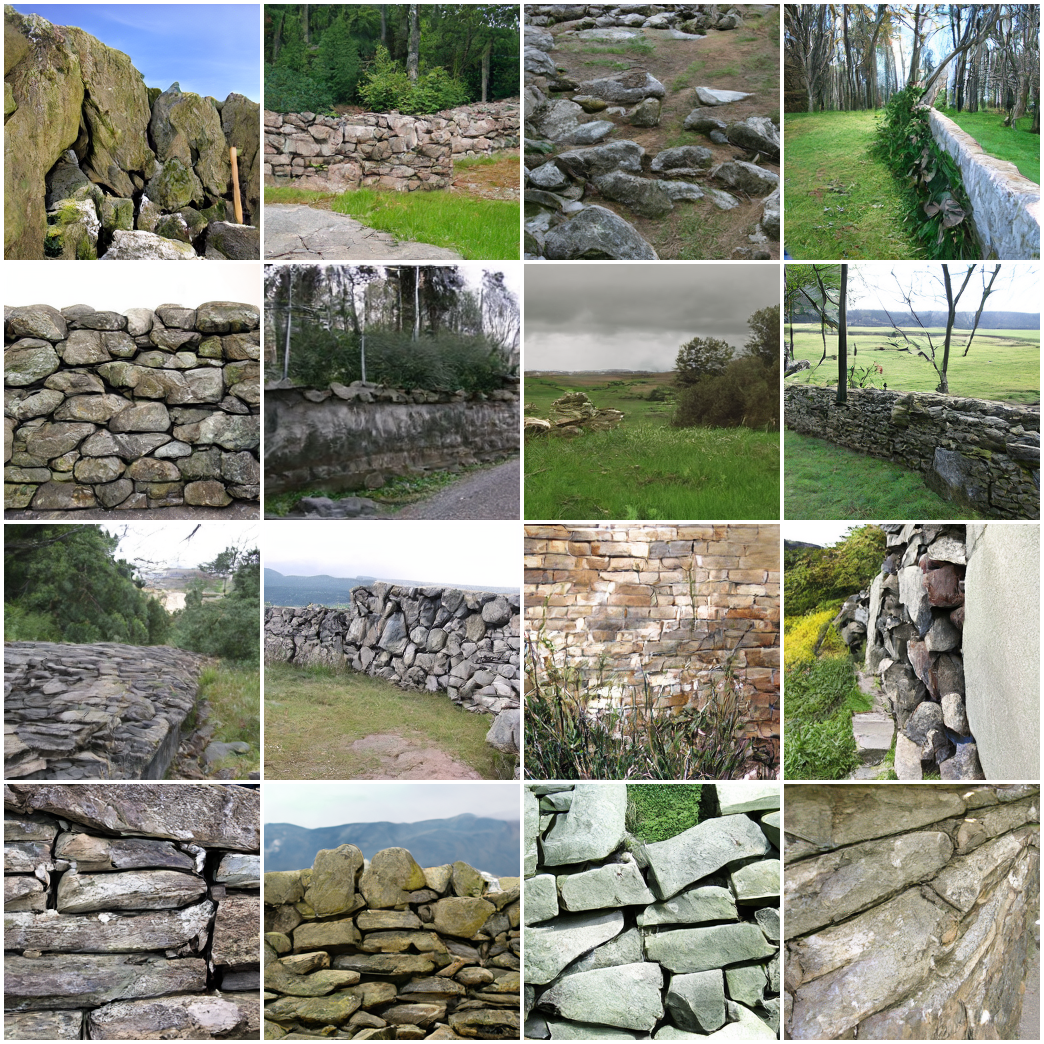}
		\caption{Stone wall (825)}
	\end{subfigure}
	\hfill
	\begin{subfigure}[t]{0.26\textwidth}
		\includegraphics[width=\textwidth]{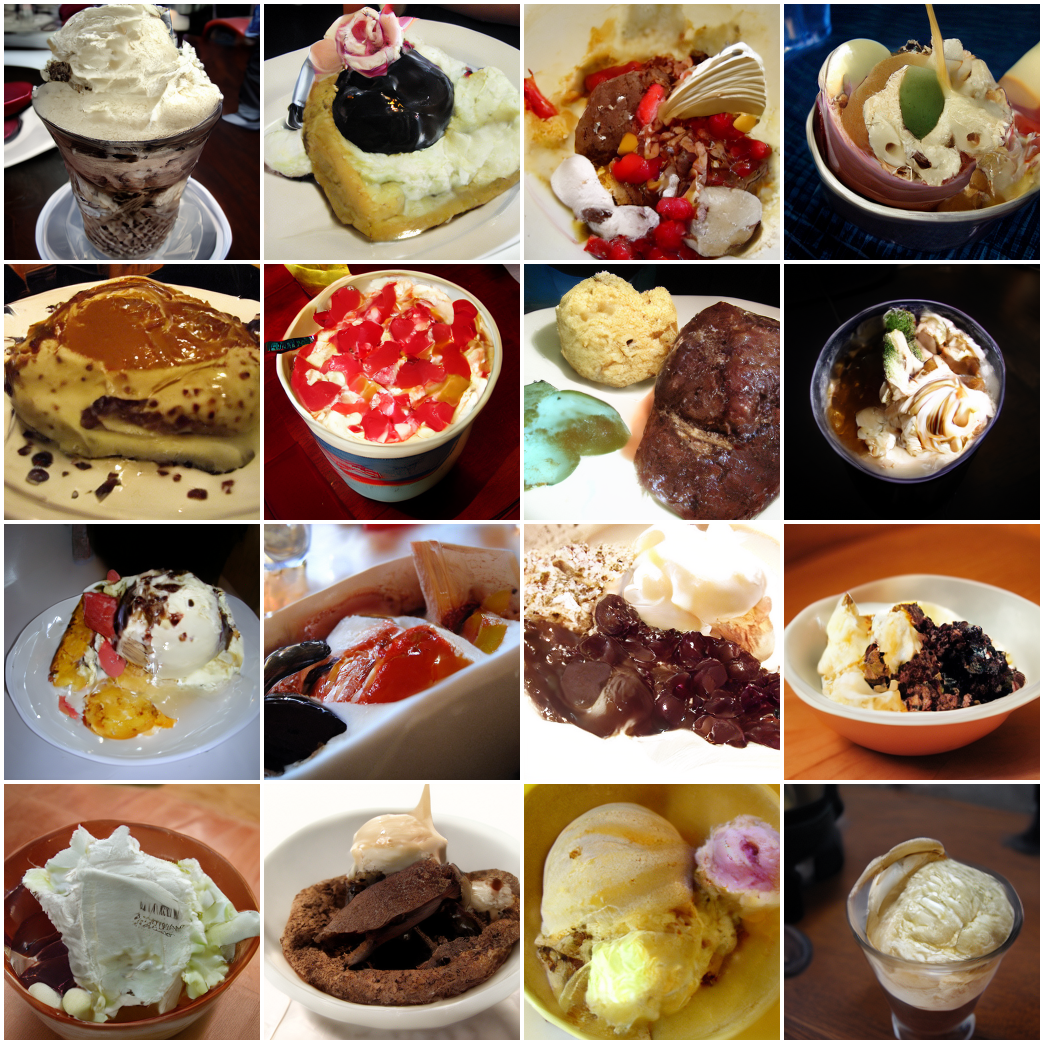}
		\caption{Ice cream (928)}
	\end{subfigure}
	\hfill
	\begin{subfigure}[t]{0.26\textwidth}
		\includegraphics[width=\textwidth]{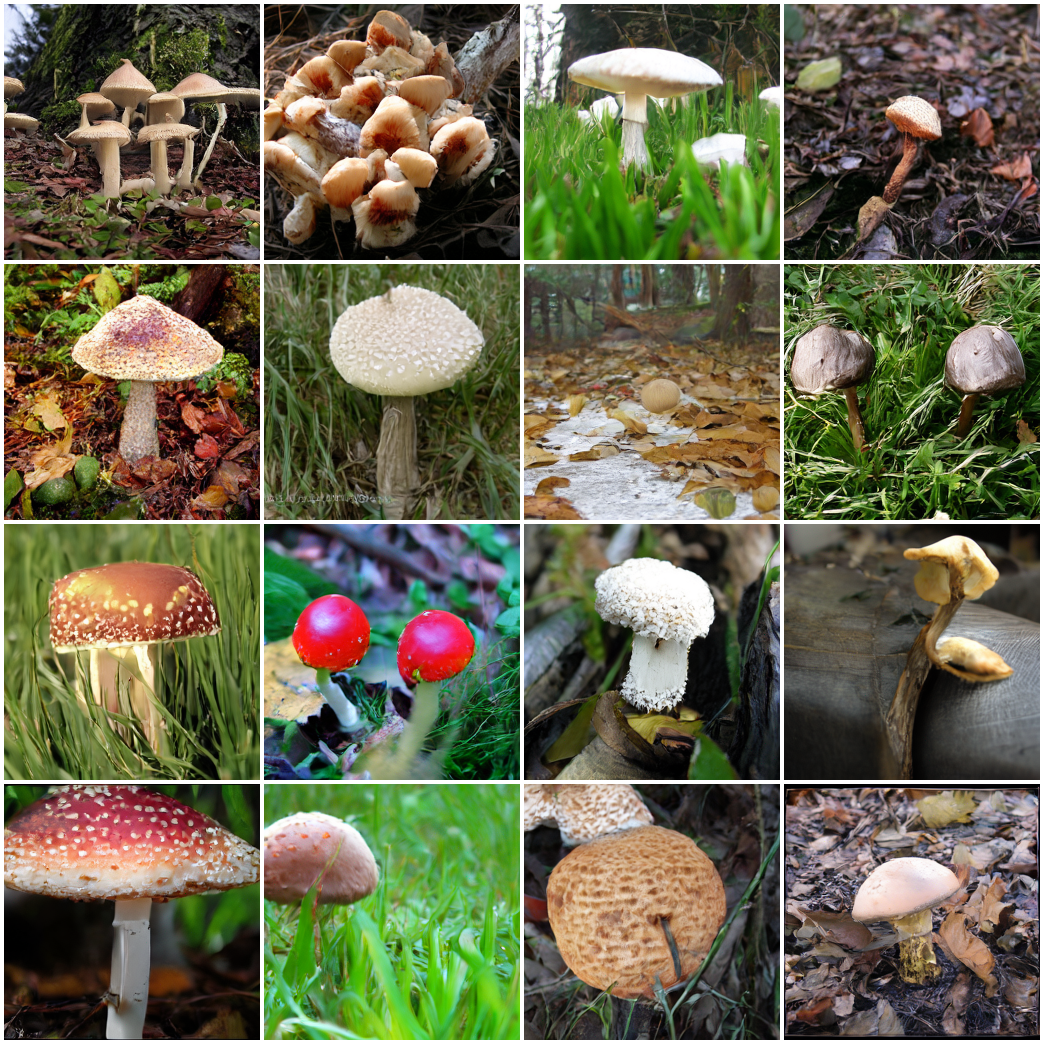}
		\caption{Mushroom (947)}
	\end{subfigure}

	\caption{Class-conditional samples from spherical-slerp SiT-XL/2 with the FLUX.2 tokenizer. Each panel: 16 samples ($4{\times}4$) for one ImageNet-1k class.}
	\label{fig:class-samples}
\end{figure}

\Cref{fig:class-samples} shows class-conditional samples from the spherical-slerp SiT-XL/2 model with the FLUX.2 tokenizer. Each panel is a $4{\times}4$ grid of 16 samples for a single ImageNet-1k class.

\subsection{Same-Class Component Swaps}
\label{sec:component-swap-qualitative}

\begin{figure}[h]
	\centering
	\includegraphics[width=0.85\textwidth]{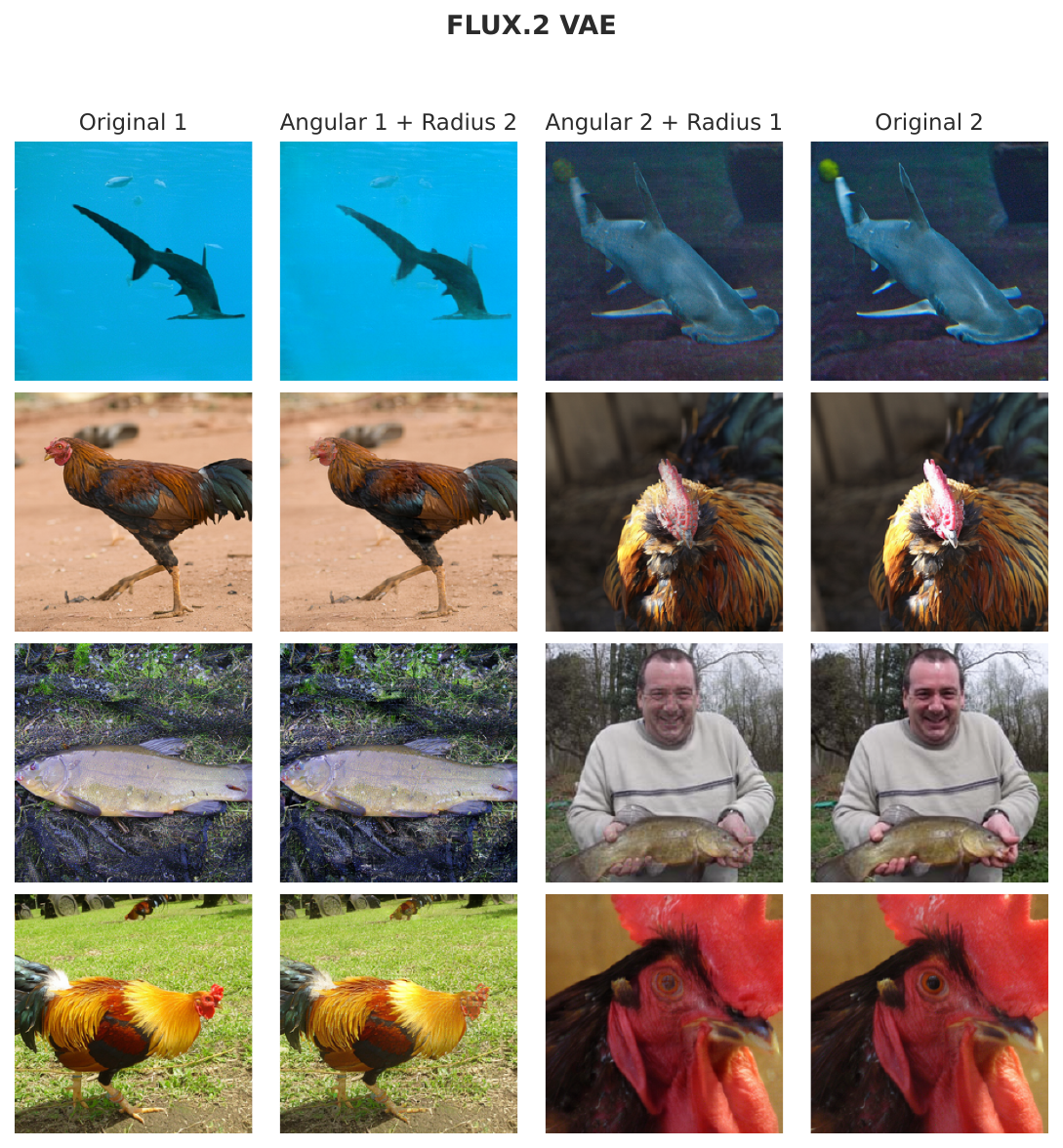}
	\caption{FLUX.2 same-class component swaps. Columns: anchor (Original 1), keep-direction hybrid (Angular 1 + Radius 2), keep-radius hybrid (Angular 2 + Radius 1), neighbor (Original 2). Each row is one (anchor, neighbor) pair.}
	\label{fig:swap-grid-flux2}
\end{figure}

\begin{figure}[h]
	\centering
	\includegraphics[width=0.85\textwidth]{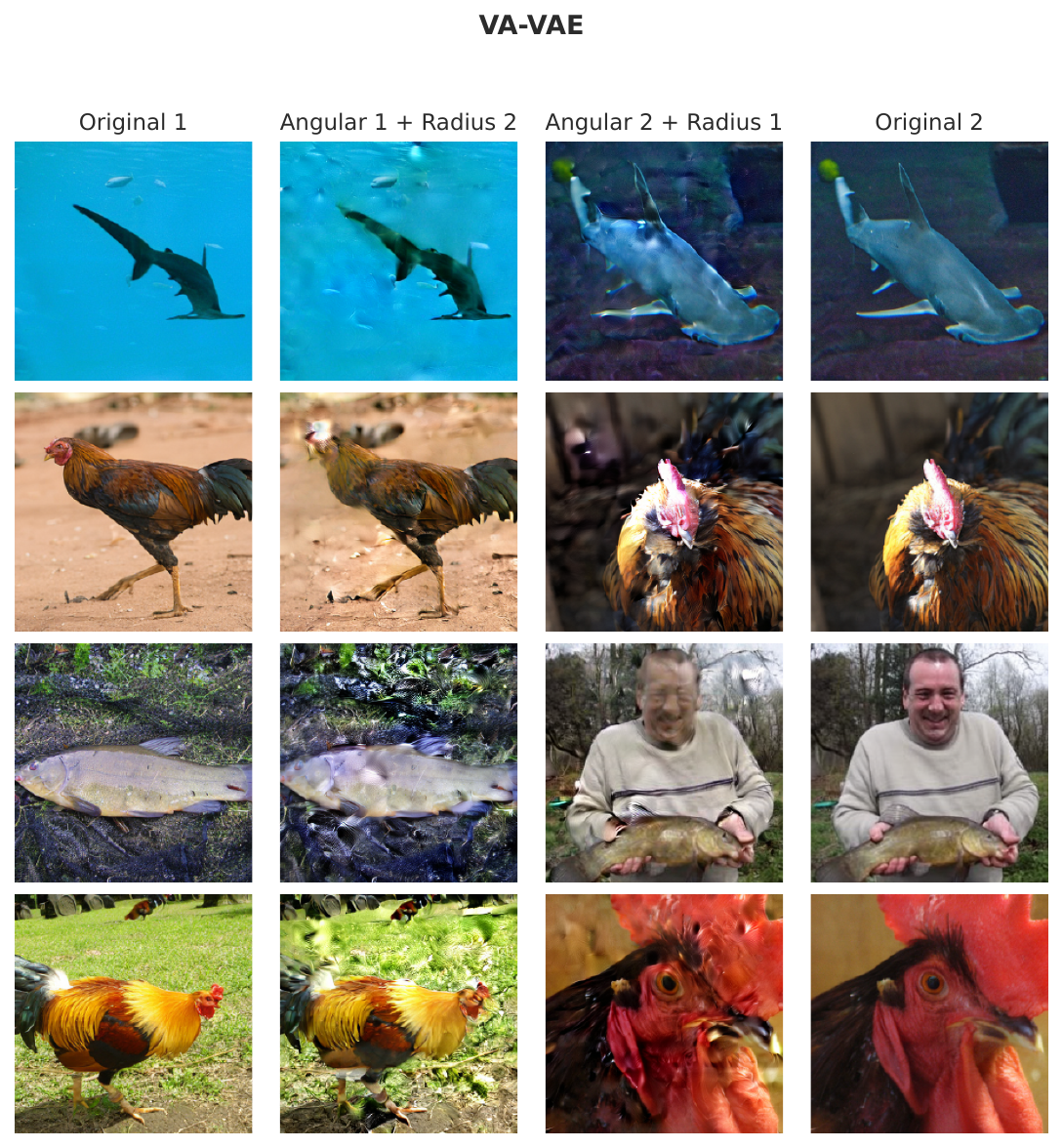}
	\caption{VA-VAE same-class component swaps; columns as in \cref{fig:swap-grid-flux2}.}
	\label{fig:swap-grid-vavae}
\end{figure}

\begin{figure}[h]
	\centering
	\includegraphics[width=0.85\textwidth]{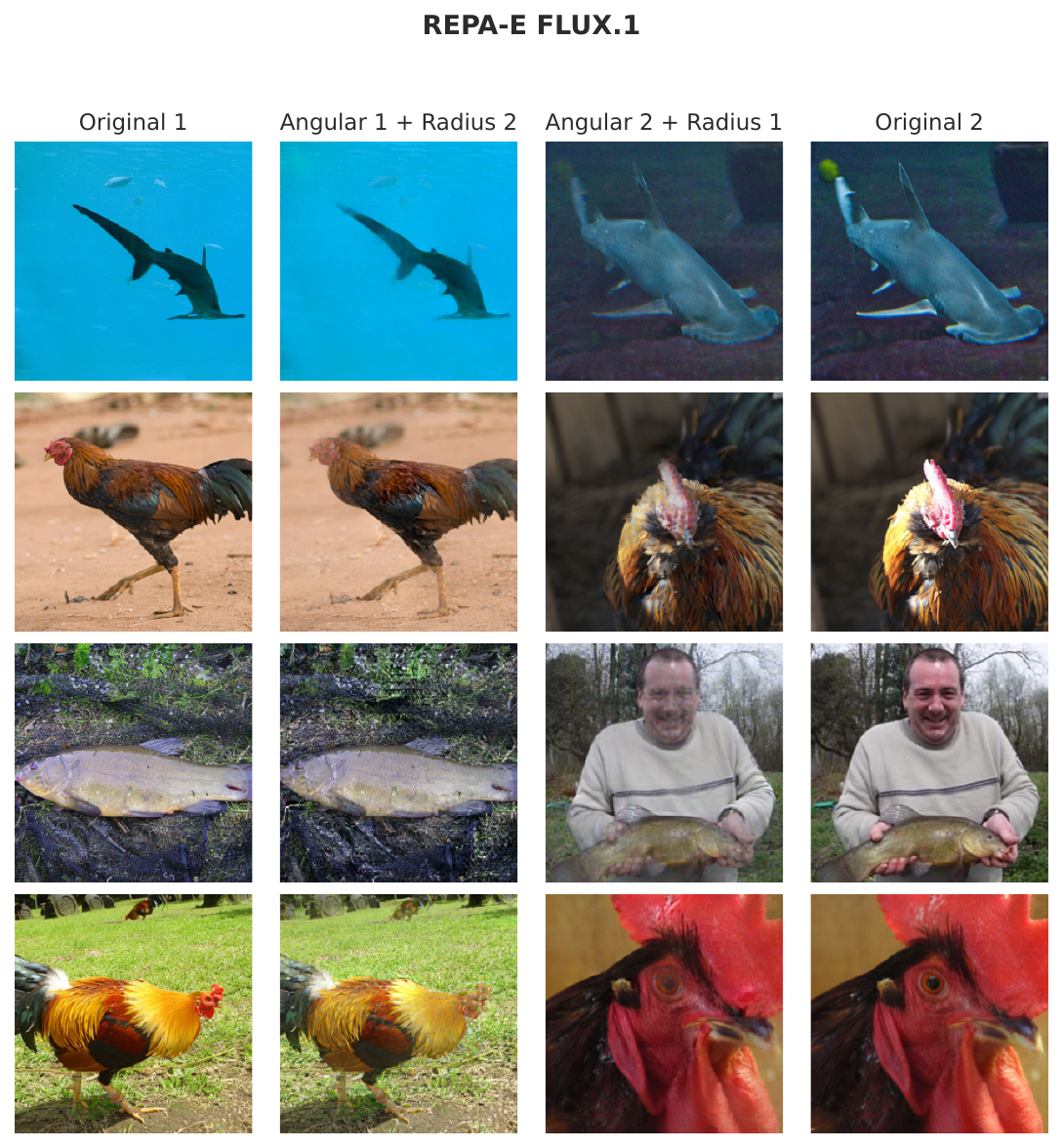}
	\caption{REPA-E FLUX.1 same-class component swaps; columns as in \cref{fig:swap-grid-flux2}.}
	\label{fig:swap-grid-repae}
\end{figure}

\Cref{fig:component-ablation} averages LPIPS and DINOv2 distance over 1024 anchor-neighbor pairs. \Cref{fig:swap-grid-flux2,fig:swap-grid-vavae,fig:swap-grid-repae} show representative pairs for each tokenizer. Columns are \emph{Original 1} (anchor), \emph{Angular 1 + Radius 2} (anchor direction with neighbor radius, the keep-direction hybrid), \emph{Angular 2 + Radius 1} (anchor radius with neighbor direction, the keep-radius hybrid), and \emph{Original 2} (neighbor). Across all three tokenizers, the keep-direction column is visually nearly indistinguishable from the anchor, while the keep-radius column resembles the neighbor.

\subsection{Reconstructions}
\label{sec:reconstructions}

\begin{figure}[h]
	\centering
	\includegraphics[width=0.55\textwidth]{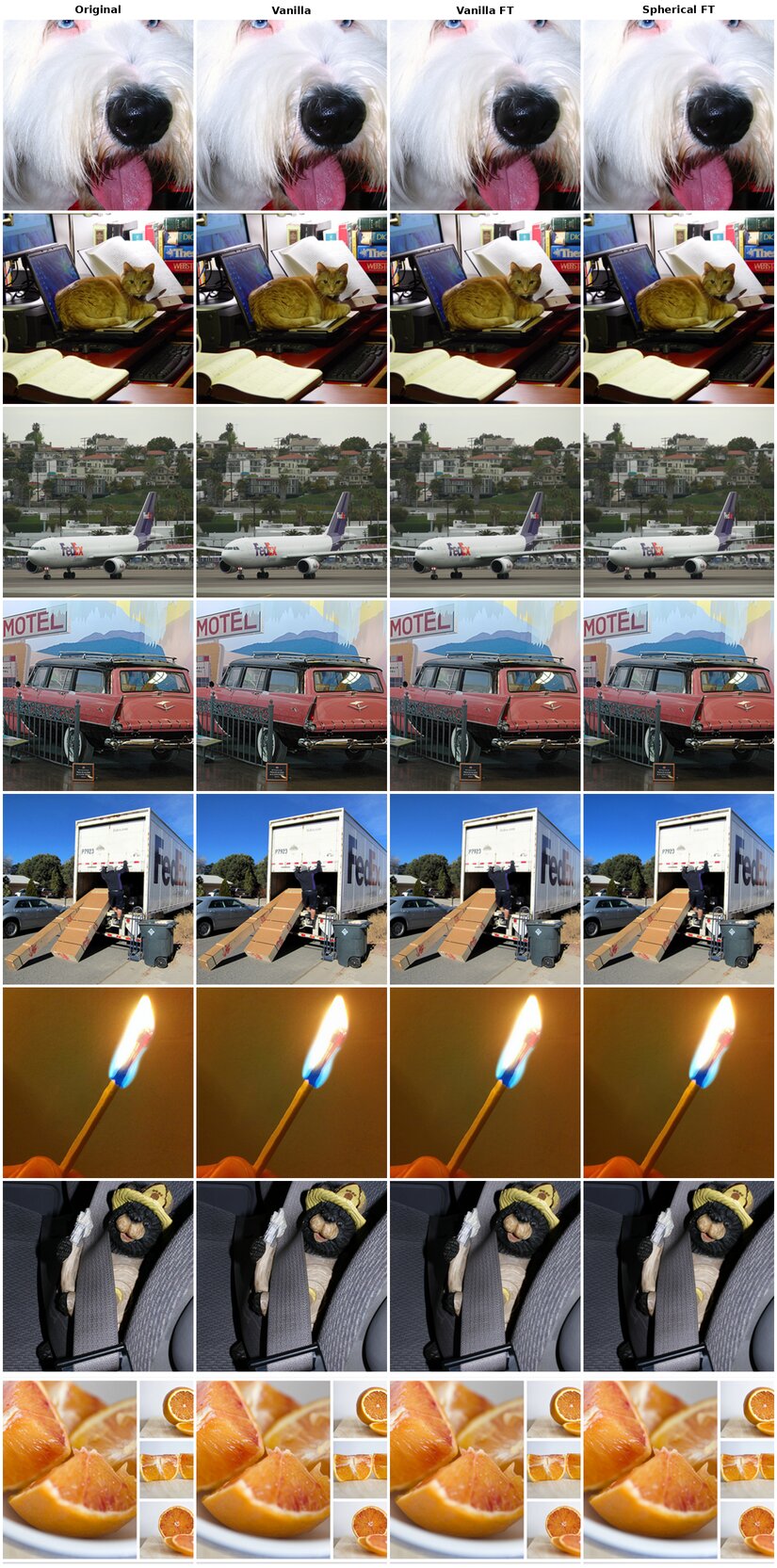}
	\caption{FLUX.2 reconstructions. Columns: original image, original (Vanilla) decoder, matched-compute vanilla decoder finetune (Vanilla FT), and spherical decoder finetune (Spherical FT). Rows: sheepdog, tabby, airliner, beach wagon, crate, matchstick, seat belt, orange.}
	\label{fig:recon-flux2}
\end{figure}

\begin{figure}[h]
	\centering
	\includegraphics[width=0.55\textwidth]{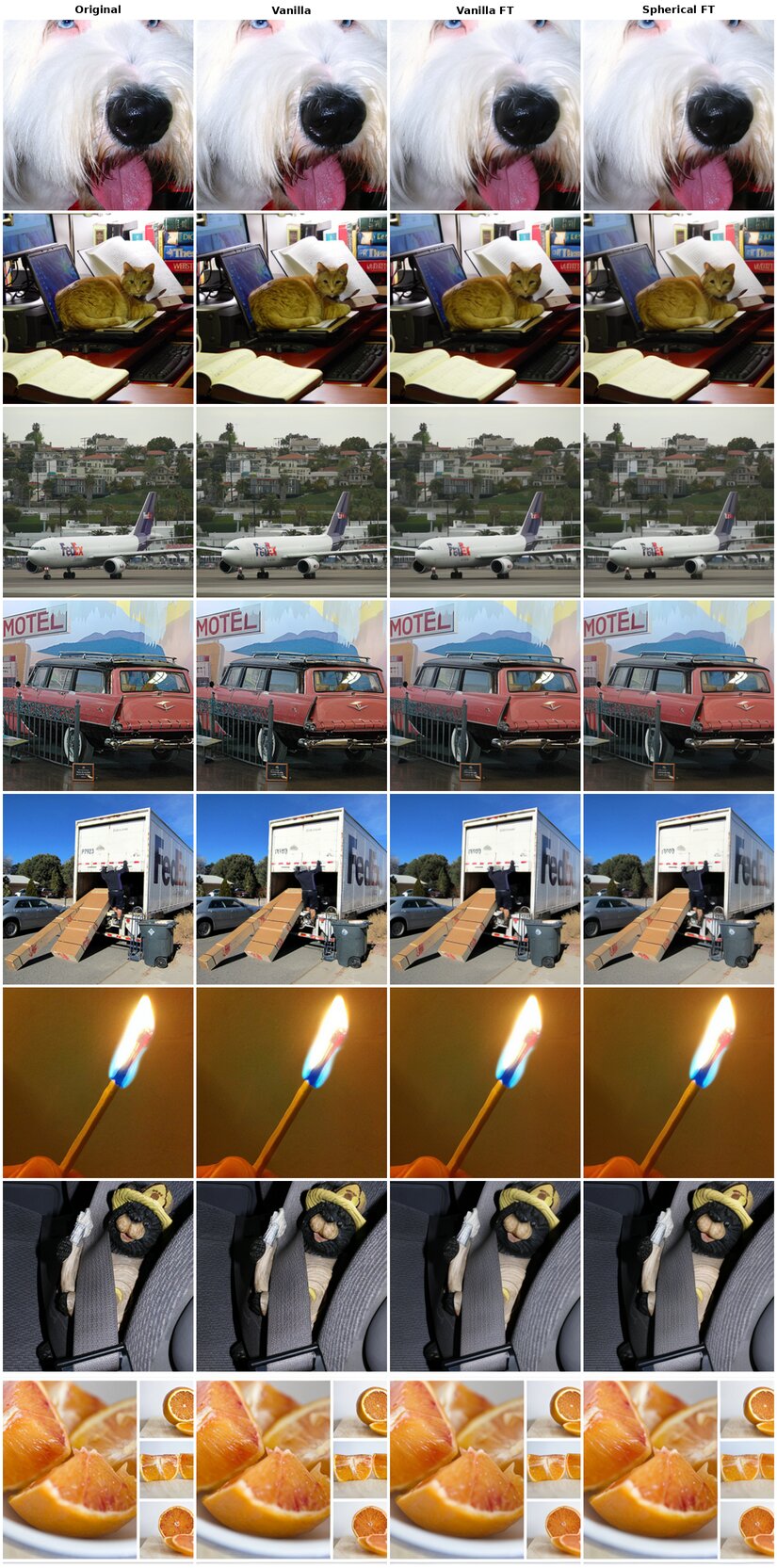}
	\caption{VA-VAE reconstructions; columns and rows as in \cref{fig:recon-flux2}.}
	\label{fig:recon-vavae}
\end{figure}

\begin{figure}[h]
	\centering
	\includegraphics[width=0.55\textwidth]{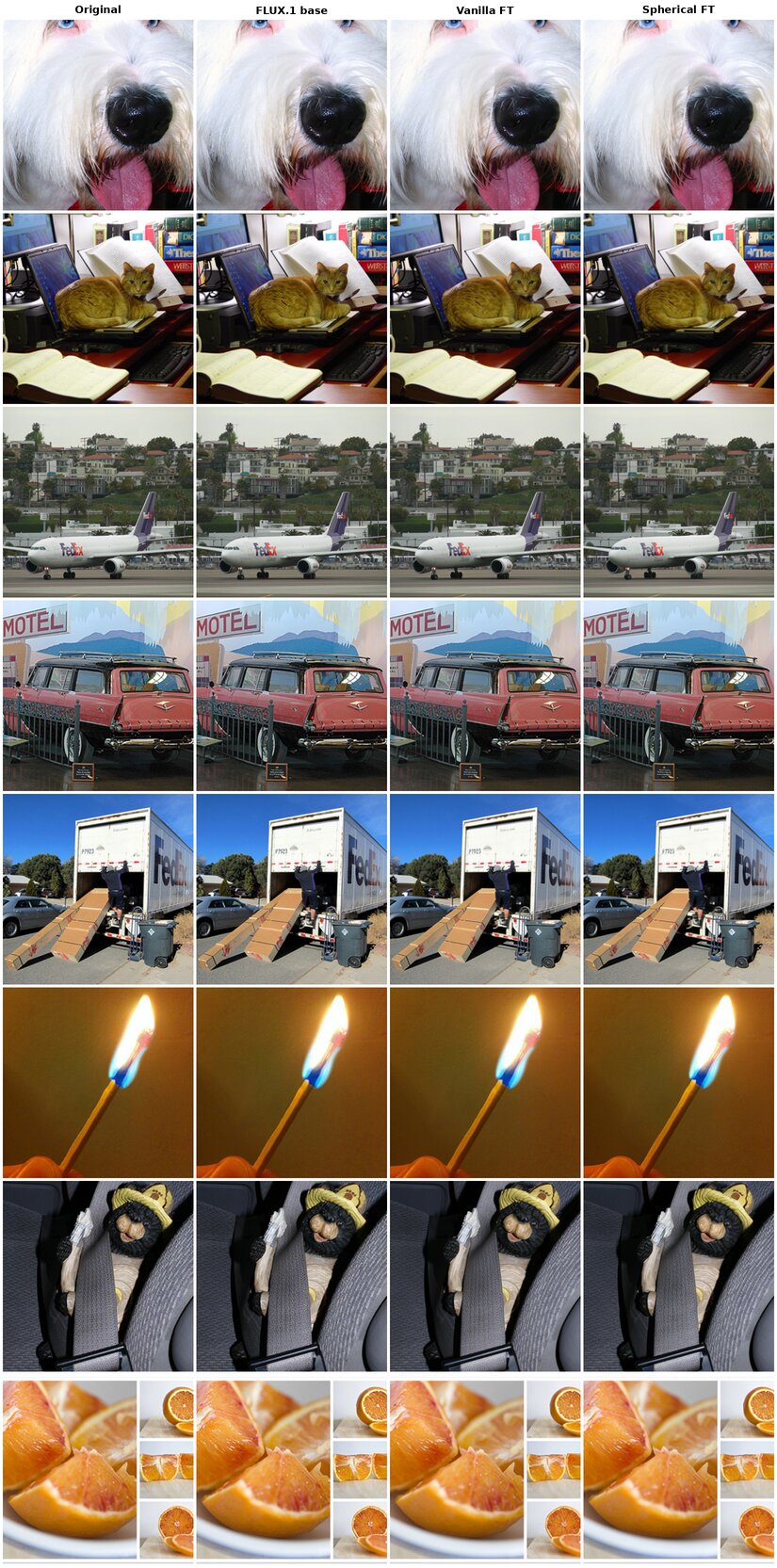}
	\caption{REPA-E FLUX.1 reconstructions; columns and rows as in \cref{fig:recon-flux2}.}
	\label{fig:recon-repae}
\end{figure}

The rFID gaps in \cref{tab:rfid} are not visible to inspection. \Cref{fig:recon-flux2,fig:recon-vavae,fig:recon-repae} compare reconstructions from the original tokenizer (Vanilla), the matched-compute vanilla decoder finetune (Vanilla FT), and the spherical decoder finetune (Spherical FT) on eight ImageNet classes. At $256{\times}256$ per tile, the three reconstructions are visually near-identical, as expected for high-fidelity tokenizers.

\section{Scope and Future Directions}
\label{sec:scope}
This work focuses on isolating the effect of latent geometry in a controlled class-conditional ImageNet-256 setting. We therefore keep the diffusion backbone, training budget, evaluator, and tokenizer-specific preprocessing fixed, and study three pretrained VAE/tokenizer families with native token dimensions $d=16$ and $d=32$. This controlled setting lets us attribute the observed gains to the spherical latent support and geodesic transport rather than to architectural or data changes. An interesting direction is to study how the same geometric constraint behaves for more compressed tokenizers, where the radial degree of freedom may play a larger role in reconstruction.

Our experiments use matched sampling budgets and guidance settings to compare latent geometries directly. A broader solver-efficiency study, including adaptive ODE solvers, lower-NFE samplers, and distillation-based acceleration, is complementary to our contribution. Since spherical-slerp keeps the latent trajectory on the fixed-radius manifold throughout sampling, future work could explore whether this structure can be exploited by specialized integrators or training-time acceleration methods.

Finally, we evaluate on ImageNet-256 to provide a clean and widely used benchmark for class-conditional generation. Extending spherical latent flow matching to text-conditioned generation, higher resolutions, non-square aspect ratios, and jointly trained encoder-decoder tokenizers would further test the generality of the approach. These directions do not require changing the central formulation: with an encoder whose outputs are projected token-wise onto a fixed-radius sphere, the flow model is trained along the corresponding spherical geodesics.

\end{document}